\documentclass[10pt,twocolumn,letterpaper]{article}

\usepackage{cvpr}              

\usepackage[dvipsnames]{xcolor}

\definecolor{cvprblue}{rgb}{0.21,0.49,0.74}
\usepackage[pagebackref,breaklinks,colorlinks,citecolor=cvprblue]{hyperref}
\usepackage{colortbl}

\def\paperID{}
\def\confName{}
\def\confYear{}

\title{Object-level Geometric Structure Preserving for Natural Image Stitching}
\author{Wenxiao Cai, \hspace{0.5em} Wankou Yang\thanks{Corresponding author: wkyang@seu.edu.cn} \\
	Southeast University\\
}

\begin{document}

\maketitle

\begin{abstract}
The topic of stitching images with globally natural structures holds paramount significance, with two main goals: pixel-level alignment and distortion prevention. The existing approaches exhibit the ability to align well, yet fall short in maintaining object structures. In this paper, we endeavour to safeguard the overall OBJect-level structures within images based on Global Similarity Prior (OBJ-GSP), on the basis of good alignment performance. 
Our approach leverages semantic segmentation models like the family of Segment Anything Model to extract the contours of any objects in a scene.
Triangular meshes are employed in image transformation to protect the overall shapes of objects within images.
The balance between alignment and distortion prevention is achieved by allowing the object meshes to strike a balance between similarity and projective transformation.
We also demonstrate that object-level semantic information is necessary in low-altitude aerial image stitching.
Additionally, we propose StitchBench, the largest image stitching benchmark with most diverse scenarios.
Extensive experimental results demonstrate that OBJ-GSP outperforms existing methods in both pixel alignment and shape preservation.
Code and dataset is publicly available at \url{https://github.com/RussRobin/OBJ-GSP}.
\end{abstract}

\section{Introduction}
\label{sec:intro}
Image stitching aims to align multiple images and create a composite image with a larger field of view. This method is widely utilized across diverse domains, including smartphone panoramic photography~\cite{mm42}, robotic navigation~\cite{mm7}, and virtual reality~\cite{mm1,mm18}. 
In recent years, the problem of alignment has largely been addressed. 
Methods such as APAP~\cite{apap} and GSP~\cite{gsp} divide the images into multiple grids, compute local transformation matrices within each grid, and combine them with global transformation information to achieve precise alignment in overlapping regions. Thus, the main concern of image stitching nowadays is to prevent distortion on the basis of good alignment performance.

\begin{figure}[t!]
	\begin{center}
    \includegraphics[width=1\linewidth]{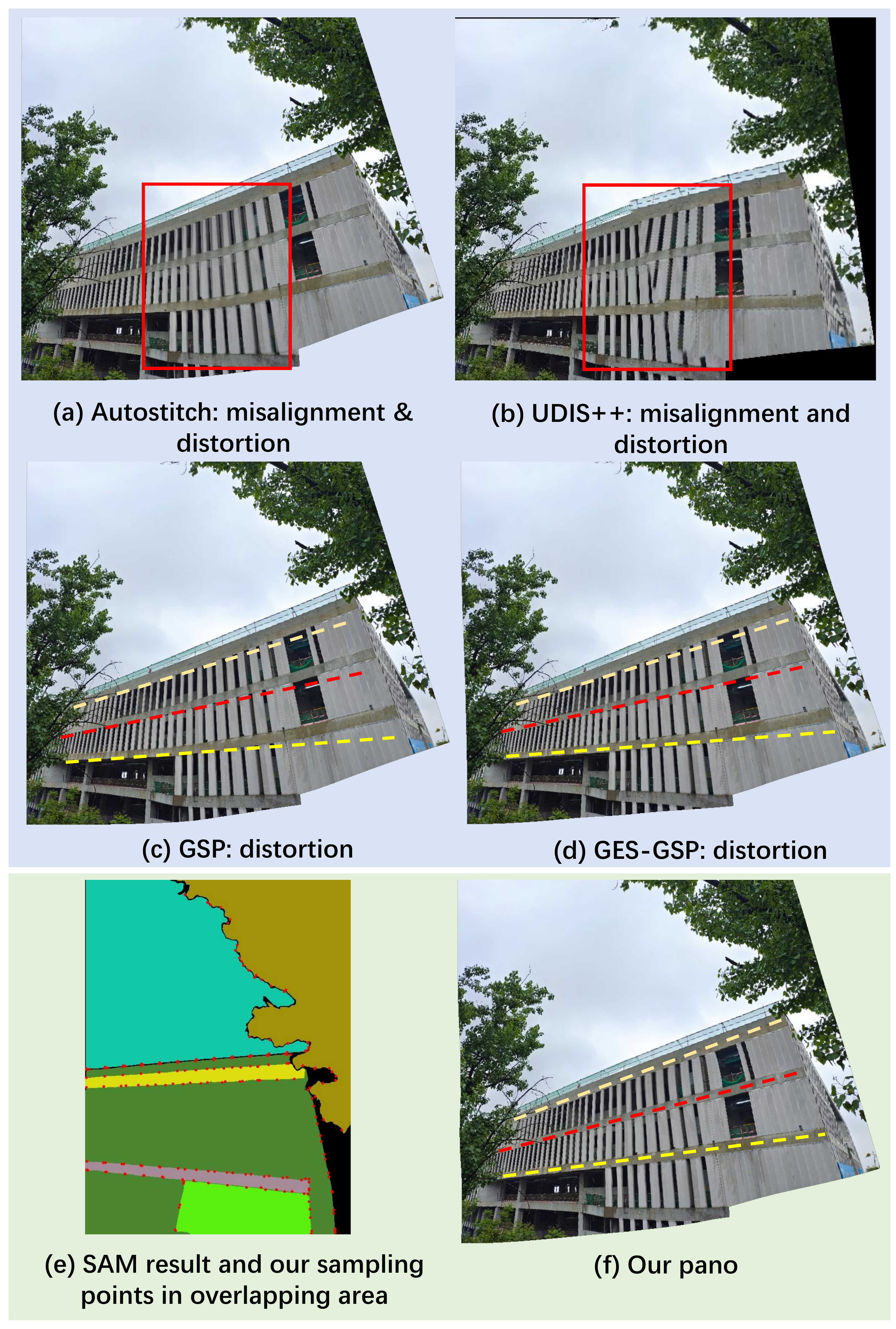}
	\end{center}
 	\caption{Red boxes indicate blurriness. (a) and (b) are not aligned well. (c) GSP~\cite{gsp} aligns well but distorts the building. Based on this, (d) GES-GSP~\cite{gesgsp} tries to prevent distortion but still fails in this case. (f) our method protects the structure of the building by sampling on object contours extracted by segmentation (e). }
	\label{fig_ges11}
\end{figure}

\begin{figure*}[t]
	\begin{center}
    \includegraphics[width=1\linewidth]{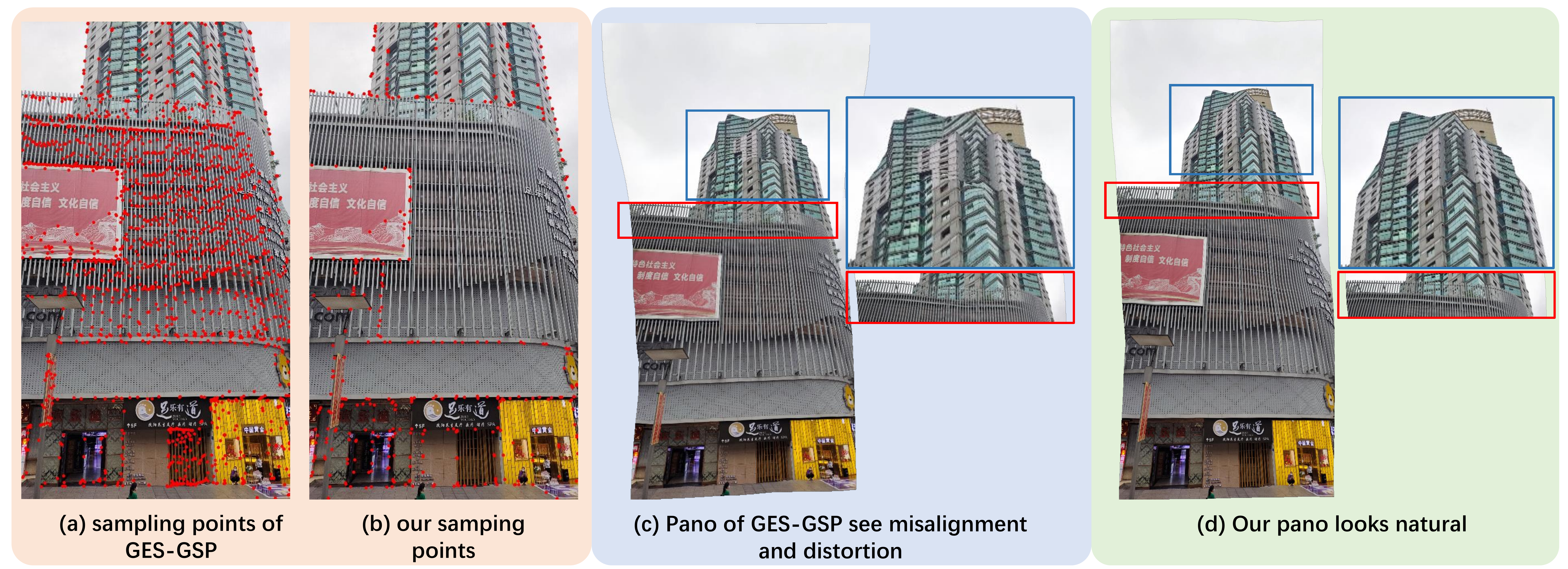}
	\end{center}
 \vspace{-1em}
 	\caption{Sampling points in our OBJ-GSP focus more on main structures so we can stitch precisely, as shown in (b)(d).}
	\label{fig_ges2}
  \vspace{-1em}
\end{figure*}

Existing works extract lines in images are preserve them in image transformation. LPC~\cite{lpc} extracts and matches lines in alignment. Based on good alignment performance of GSP, GES-GSP~\cite{gesgsp} adds the similarity transformation of line structures into considerations. 
However, (a) they only preserves line structures, ignoring overall and object-level structures, (b) focusing only on individual lines can be quite chaotic and mislead the model (Fig.~\ref{fig_ges2}), (c) straight or curved structures do not exist in some scenes.


Since an important criterion for humans to judge whether an image looks natural is the naturalness of the object structures within the image, our key insight is to extract these structures and preserve them during stitching.
Nowadays, state-of-the-art segmentation models can identify almost any object with superior performance. We use them to get object shapes, which represents the image structure, and then use triangle meshes to preserve these segmented object shapes during the stitching.
We generate triangle meshes within each object. During image transformation, these triangle meshes tend to reach a balance between projection and similarity transformations, effectively preserving the structure of the objects. 
As demonstrated in Fig.~\ref{fig_ges11} (f), our method excels in maintaining the overall structure of images by preventing distortion of prominent object shapes. 
OBJ-GSP capitalizes on object-level preserving, and we adopt leverage segmentation models to extract geometric information.
As shown in Fig.~\ref{fig_ges11} (e), segmentation models treats objects as cohesive entities, transcending the segmentation of individual lines and curves adopted in previous works~\cite{gesgsp,lpc}. 
This allows for a more nuanced understanding of the relationships between individual geometric structures, and superior to previous work, it works even when there are no prominent linear structures in the images.

Previous works often used their own collected images without testing on datasets from other papers. We unified the datasets from previous works and incorporated our own collected hand-held camera and aerial images to create StitchBench, the most complete benchmark to date.
We also demonstrate that in low-altitude aerial image stitching, semantic segmentation in OBJ-GSP pipeline is necessary. When the drone flies at a low altitude, the camera moves significantly, and there is a considerable distance difference between the roofs and the ground relative to the camera. 
These conditions do not satisfy the assumptions of image stitching~\cite{autostitch}, which assumes a fixed camera optical center or distant scenes, making stitching unfeasible. In this case, it is necessary to use a semantic segmentation model to identify the houses, then perform orthorectification to project them onto the ground before stitching.

To summarize, the main contributions of the proposed OBJ-GSP include:
\begin{itemize}
    \item We propose to preserve object contours before and after image transformation to maintain the overall structure of the image. Object shapes are not limited to images with obvious linear structures and are not misled by excessively noisy line structures.
    \item We introduce the segmentation models into image stitching, facilitating the extraction of any object in the scene. Furthermore, we demonstrate that segmentation and OBJ-GSP are crucial for low-altitude aerial image stitching.
    \item We collect StitchBench, which is by far the largest and most diverse image stitching benchmark.
\end{itemize}

\section{Related work}
\label{sec:relatedworks}

\subsection{Grid-based image stitching}
Autostitch \cite{autostitch}, a pioneering work in image stitching, matches feature points and aligns them by homography transformation. Building upon this foundation, numerous stitching algorithms partition images into grids, compute geometric transformation relationships for each grid, and combine them into a global transformation to align overlapping regions and seamlessly transit the transformation to non-overlapping areas.
APAP~\cite{apap}, AANAP~\cite{aanap}, and GSP~\cite{gsp} have evolved over time, essentially addressing most alignment problems in images.
However, their grid deformation methods have no knowledge of object shapes. They pay too much attention on alignment and thus causes geometric distortion. 
To address this, LPC~\cite{lpc} and GES-GSP~\cite{gesgsp} propose to preserve line structures.
However, 
(a) their method only preserves line structures, ignoring the overall structure of objects, 
(b) an excessive number of lines without object structure information can mislead the model, 
(c) some scenes do not contain straight or curved structures.
We find that the large segmentation models like SAM~\cite{sam} can segment all types of objects and provide their contours. This helps image stitching maintain shape consistency, so we have incorporated the family of SAM into our method.
We use triangular grids to protect the overall object-level geometric structure, and establish connections between dispersed geometric transformations, achieving superior results.

\subsection{Geometric structure extraction}
Previous works employ Line Segment Detector~\cite{lsd} to detect straight lines in images, and edge detection methods like Canny~\cite{canny} and HED~\cite{hed} to identify edges. However, these methods require line structures to be present in the image.
In cases where textures are unclear or lighting is poor, conventional methods cannot extract lines effectively, whereas large models can still operate successfully in these scenarios.
We employ the family of SAM and EfficientSAM~\cite{efficientsam} to extract object-level structures and preserve them during stitching. 
It is notable that segmentation models are not limited by line structures and can segment almost any object.
In the future, the accuracy and speed of SAM-type methods will both improve~\cite{efficientsam,lightweightsam}, further enhancing the quality and speed of our image stitching techniques.

\subsection{Deep-learning based stitching}
In recent years, several methods~\cite{deep1} like UDIS~\cite{udis} have attempted to model certain image stitching steps as unsupervised deep learning problems, leading to notable advances in this field.
UDIS++~\cite{udisplus} also addresses the distortion problem on the basis of good alignment performance, which aligns perfectly with our goals. 
We adhere to the traditional approach in the stitching domain by preserving results in grid transformation, while UDIS++ provides a completely new deep learning-based pipeline, although currently its performance is not as good as ours. 

\section{The proposed method}
\label{sec:proposedmethod}

\begin{figure}[h]
	\begin{center}
    \includegraphics[width=1\linewidth]{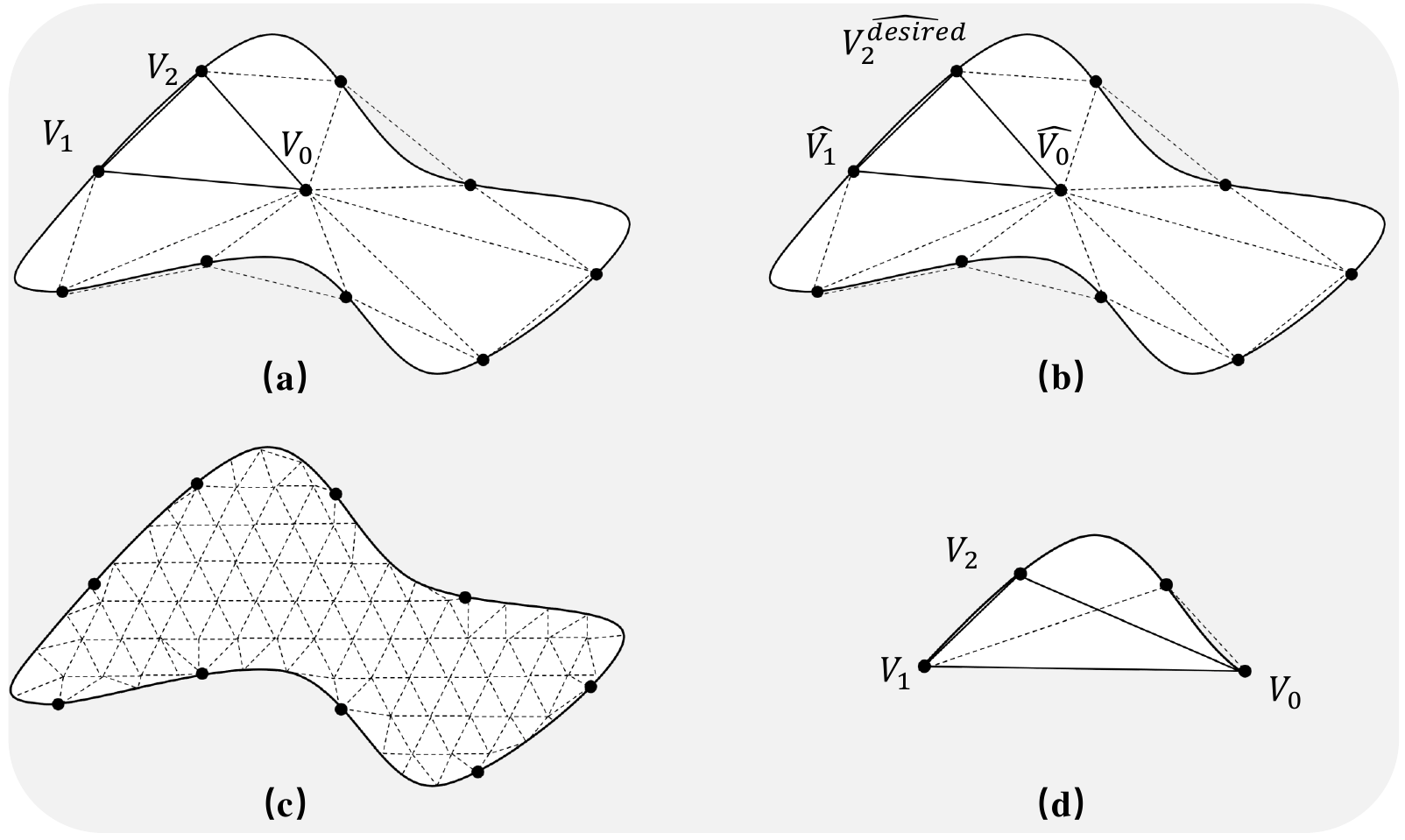}
	\end{center}
 	\caption{(a) Our triangle mesh. $V_0$ is the center of object. (b) With every edge undergoing similarity and projection transformation, object in (a) is transformed into (b). (c) Triangle mesh with  near-equilateral triangles of similar sizes across the region. (d) Triangle sampling strategy.}
	\label{fig_mesh}
\end{figure}

OBJ-GSP introduces SAM to segment objects to obtain their structural contours and preserve object-level structures as well as aligning feature points in stitching. 
Locally, our approach retains the original perspective of each image. On a global scale, it seeks to preserve overall structure~\cite{gsp}. 
Moreover, at the object-level, we ensure the integrity of objects within the images, preventing distortion.
To this end, we take four aspects into consideration: alignment, global similarity, local similarity and object-level shape preservation.
A grid mesh is adopted to guide the image deformation, where $V$ and $E$ represent the sets of vertices and edges within the grid mesh, as shown in Fig.~\ref{fig_mesh}. Image stitching methods aim to find a set of deformed vertex positions, denoted as $\widetilde{V}$, that minimizes the energy function $\psi(V)$.

\textbf{Alignment term} extracts feature points $p$ by with an extractor (e.g. SIFT~\cite{sift}) and matches feature point pairs with matcher $\Phi$. 
For each feature point pair $(p, \Phi(p))$, $\widetilde{v}(p)$ represents the position of $p$ as a linear combination of four vertex positions, and $M$ represents the set of all feature point pairs. The algorithm linearly combines the coordinates of the four vertices of each grid to represent the position of $p$ through bi-linear interpolation. By optimizing the positions of grid vertices after geometric transformation, it aims to bring $p$ as close as possible to $\Phi(p)$. Therefore, the energy equation is defined as:

\begin{equation}
\psi_a(V) =  \sum_{p_k \in M} \|   \widetilde{v}(p_k) - \widetilde{v}(\Phi(p_k)) \|^2.
\end{equation}

\textbf{Local similarity term} aims to ensure that the transition from overlapping to non-overlapping regions is natural. Each grid undergoes a similarity transformation to minimize shape distortion. For an edge $(j, k)$, $S_{jk}$ represents its similarity transformation. Suppose $v_j$ transforms to $\widetilde{v_j}$ after deformation, and the energy function is defined as:

\begin{equation}
\psi_l(V) = \sum_{(j,k) \in E_i} \| (\widetilde v_k - \widetilde v_j) - S_{jk}(v_k - v_j) \|^2.
\end{equation}

\textbf{Global similarity term} operates on a global scale to ensure the entire image undergoes a similarity transformation. GSP algorithm evaluates the scale $s$ and rotation $\theta$ within the global image transformation and computes parameters $c(e)$ and $s(e)$ for similarity. Thus, the energy function is defined as:

\begin{equation}
\psi_g(V) = \sum_{e_j \in E} w(e_j)^2[(c(e_j) - s \cos\theta)^2 + (s(e_j) - s \sin\theta)^2].
\end{equation}

\begin{figure*}[h]
	\begin{center}
    \includegraphics[width=1\linewidth]{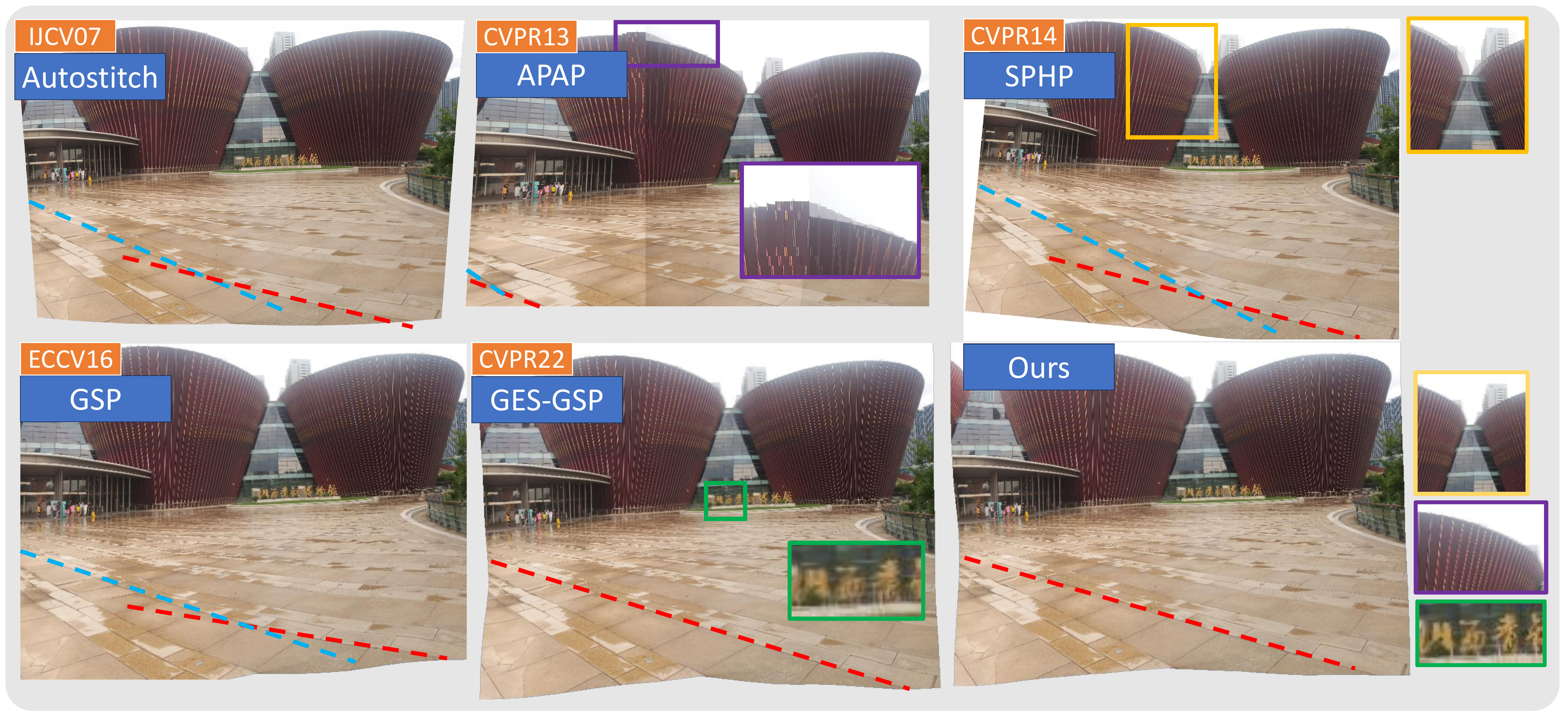}
	\end{center}
 	\caption{APAP and SPHP see misalignment, and we delineated the indistinct portions using color-coded boxes. Autostitch, APAP, SPHP and GSP sees distortion. The convergence of the blue and red lines is essential, and we signify distortion by the intersection of these two lines. GES-GSP successfully prevents distortion, but it undergoes misalignment. Our method addresses misalignment and distortion well.}
	\label{fig_ges21}
\end{figure*}

After obtaining the contours, we generate a triangular mesh for each semantic object, preserving the shape of the object through similarity transformations within the triangle mesh. Unlike the As-Rigid-As-Possible (ARAP) \cite{arap} method, we simplify computational complexity by directly locating the center of the object and connecting it to sampling points on the object's semantic boundary to form a triangular mesh. In Fig \ref{fig_mesh}, $V_0$ represents the object's center, while $V_1$ and $V_2$ are sampling points on the semantic boundary of the object, forming a triangular mesh with these three points. $(x_{01},y_{01})$ refer to the known coordinates of a feature point in the local coordinate plane. One vertex, $V_2$, of the triangle can be represented using the edges of the triangle $\overrightarrow{V_0V_1}$ and an orthogonal coordinate system obtained by rotating this edge counterclockwise by 90 degrees:

\begin{equation}
V_2 = V_0 + x_{01}\overrightarrow{V_0V_1} + y_{01}\left[\begin{matrix}0&1\\-1&0\\\end{matrix}\right]\overrightarrow{V_0V_1}.
\end{equation}

\begin{figure*}[h]
	\begin{center}
    \includegraphics[width=1\linewidth]{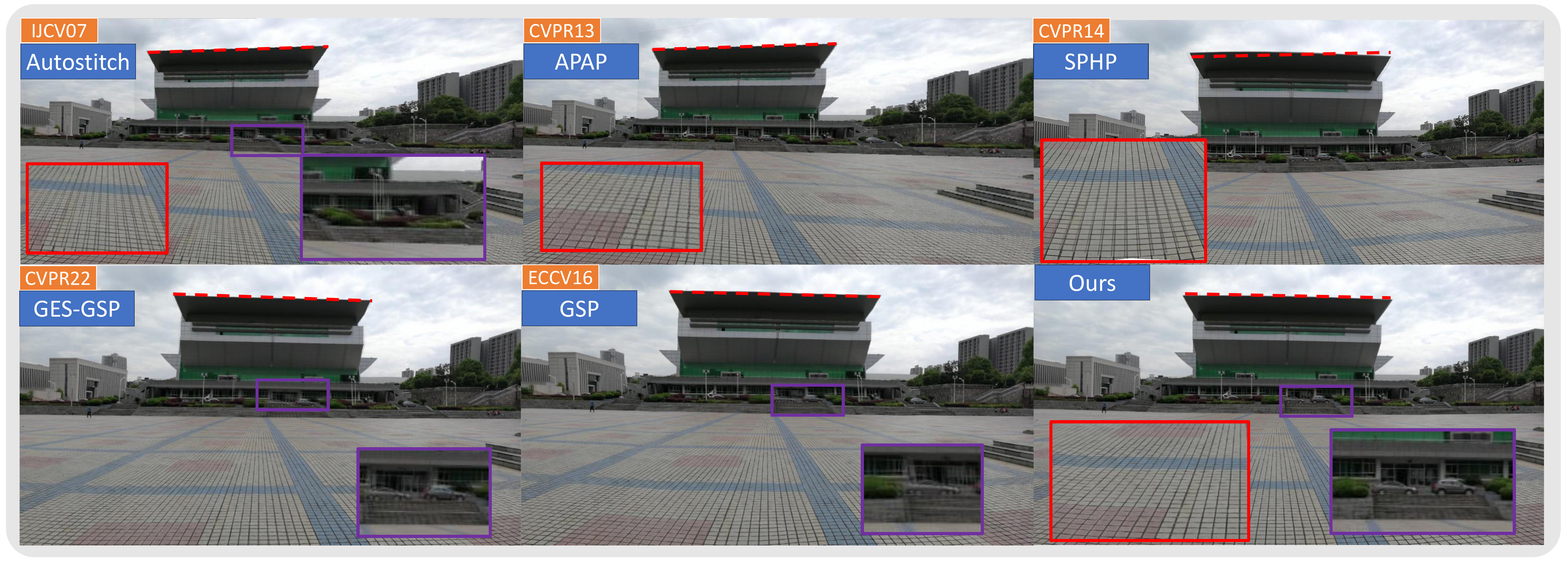}
	\end{center}
 	\caption{We magnify the ground in the red box and the car in the purple box. OBJ-GSP aligns well but the remaining five methods shows misalignment. Distortion is not observed in this case.}
	\label{fig_rewgym}
\end{figure*}

After the mesh deformation, $V_0$ and $V_1$ are transformed into $\widehat{V_0}$ and $\widehat{V_1}$. To preserve the shape of the segmentation result, we aim for the triangle to undergo a similarity transformation, keeping $x_{01}$ and $y_{01}$ unchanged. Therefore, we desire $V_2$ to transform into:

\begin{equation}
\widehat{V_2^{desired}} = \widehat{V_0} + x_{01}\overrightarrow{\widehat{V_0}\widehat{V_1}} + y_{01}\left[\begin{matrix}0&1\\-1&0\\\end{matrix}\right]\overrightarrow{\widehat{V_0}\widehat{V_1}}.
\end{equation}

The corresponding energy term for the transformed $\widehat{V_2}$ is calculated as:

\begin{equation}
E_{V_2} = \|\widehat{V_2^{desired}} - \widehat{V_2}\|^2.
\end{equation}

Similar definitions for energy terms are applied to $\widehat{V_0}$ and $\widehat{V_1}$, resulting in the error sum for a triangle:

\begin{equation}
E_{\{V_0,V_1,V_2\}} = \sum_{i=0,1,2} \|\widehat{V_i^{desired}} - \widehat{V_i}\|^2.
\end{equation}

Initially, our approach constructs the triangular mesh by selecting sampling points and the object's center. Unlike ARAP \cite{arap}, we do not employ equilateral triangular meshes, as objects segmented from the image often lead to very small equilateral triangles. Experimental results demonstrate that this approximation not only has no adverse impact on the final outcome but also reduces computational complexity:

\begin{equation}
E_{{{V_0,V_1,V}_2}} = \sum_{i=1,2} \|\widehat{V_i^{desired}} - \widehat{V_i}\|^2.
\end{equation}

We extract  $N_c$ semantic object structures from a single image using semantic segmentation, and $N_s$ represents the total number of all sampling points within geometric structure i. Similar to GES-GSP \cite{gesgsp}, $\omega$ is a coefficient calculated based on the positions of the sampling points. Consequently, the total error equation is as follows:
\begin{equation}
\psi_{obj}(V) = \sum_{\beta = 1}^{N_c}\sum_{\alpha = 1}^{N_s}{\omega_\alpha^\beta E_\alpha^\beta}.
\end{equation}

To conclude, our objective function is given by:

\begin{equation}
\label{eq_final_optimization}
\begin{split}
\widetilde{V} &= \underset{\widetilde{V}}{arg\min}\left(\psi_a(\widetilde{V}) + \lambda_l\psi_l(\widetilde{V}) \right.\\
&\quad\left. + \psi_g(\widetilde{V}) + \lambda_{obj}\psi_{obj}(\widetilde{V})\right).
\end{split}
\end{equation}

Eq.~\ref{eq_final_optimization} can be solved with linear optimization.
For fair comparison, our parameters are identical to those of GES-GSP: $\lambda_l = 0.75$, $\lambda_{obj} = 1.5$. Our $\lambda_{obj}$ corresponds $\lambda_{ges}$ in to GES-GSP.

\begin{figure}[h]
	\begin{center}
    \includegraphics[width=1\linewidth]{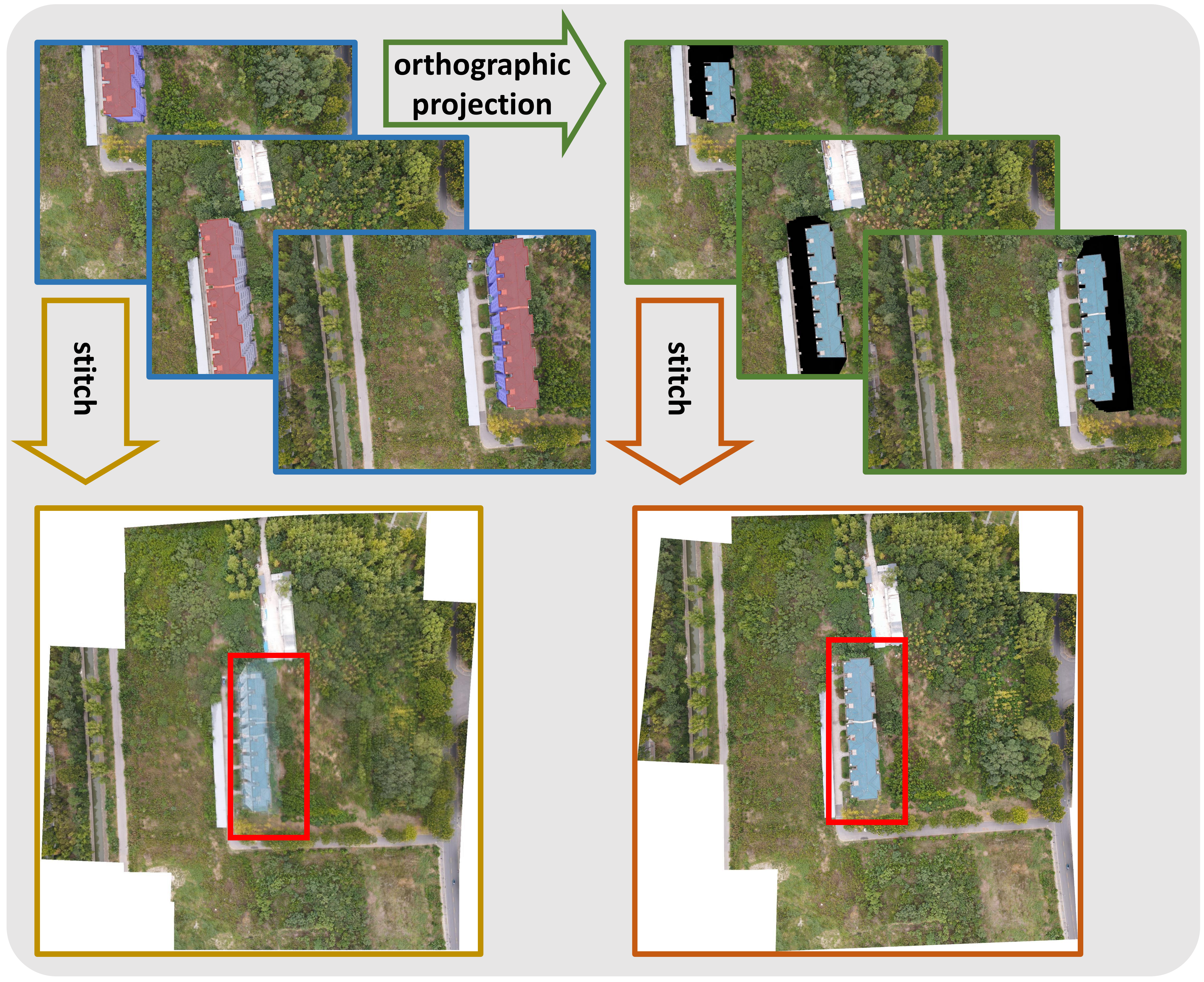}
	\end{center}
 	\caption{OBJ-GSP for low-altitude drone image stitching. We segment our roofs and walls from aerial images first. Walls are masked out, and roofs are projected to the ground. Then OBJ-GSP stitches the projected images.}
	\label{fig_aerialpipeline}
\end{figure}

\begin{table*}[h!]
    \caption{We report the mean MDR for distortion prevention evaluation, and NIQE to measure alignment performance. UDIS and UDIS++ are not feature point based so we only report NIQE and leave qualitative results in supplementary material. Best results are labeled with \textbf{bold text}. Lower MDR and NIQE indicates better stitched panorama. Improvement row compares our proposed method and GES-GSP. Our mean improvement to GES-GSP is \textbf{3.5\%} in MDR and \textbf{3.8\%} in NIQE.}
    \setlength{\tabcolsep}{15pt} 
    \begin{center}
        \begin{tabular}{l|c|c|c|c|c|c}
\toprule
            Models / Datasets & OBJ-GSP & AANAP & APAP & CAVE & DFW & DHW \\
            \hline
            \multicolumn{7}{c}{Mean Distorted Residuals (MDR  $\downarrow$)} \\
            \hline
            GSP \textcolor{blue}{\fontsize{7}{10}\selectfont ECCV16} & 1.15296 & 1.06183 & 1.25495  & 0.90884 & 0.98457 & 1.08755 \\
            GES-GSP \textcolor{blue}{\fontsize{7}{10}\selectfont CVPR22} & 1.14366 & 1.06213 & 1.24249  & 0.90821 & 0.98034 & 1.05619  \\
            \textbf{OBJ-GSP (ours)} & \textbf{1.12229} & \textbf{1.05930} & \textbf{1.20123} & \textbf{0.89731} & \textbf{0.97259} & \textbf{1.00496}\\
            \rowcolor{lightgray} Improvement(\%) & 1.9 & 0.3 & 3.3 & 1.2 & 0.8 & 4.9 \\
            \hline
            \multicolumn{7}{c}{Naturalness Image Quality Evaluator (NIQE $\downarrow$)} \\
            \hline
            UDIS \textcolor{blue}{\fontsize{7}{10}\selectfont TIP21}& 3.69421 & 3.01517 & 3.69421 & -  & 5.74137 & 3.28645 \\
            UDIS++ \textcolor{blue}{\fontsize{7}{10}\selectfont ICCV23}& 3.34003 & 2.95493 & 3.56812 & 4.07702 & \textbf{5.09680} & 3.23392 \\
            GSP \textcolor{blue}{\fontsize{7}{10}\selectfont ECCV16}& 2.66597 & 2.84241 & 3.4356  & 4.04708 & 5.61905 & 2.75485 \\
            GES-GSP \textcolor{blue}{\fontsize{7}{10}\selectfont CVPR22}& 2.64986 & 2.77220 & 3.48713  & 4.03835 & 5.71544 & 2.70838 \\
            \textbf{OBJ-GSP (ours)} & \textbf{2.54906} & \textbf{2.74965} & \textbf{3.39280} & \textbf{4.01565} & 5.69104 & \textbf{2.60825} \\
            \rowcolor{lightgray} Improvement(\%) & 3.8 & 0.8 & 2.7 & 0.6  & 0.4 & 3.7 \\
            \bottomrule

            Models / Datasets & GES-GSP & LPC & REW & SEAGULL & SVA & SPHP \\
            \hline
            \multicolumn{7}{c}{Mean Distorted Residuals (MDR $\downarrow$)} \\
            \hline
            GSP \textcolor{blue}{\fontsize{7}{10}\selectfont ECCV16}& 1.06986 & 1.30562 & 1.16192  & 1.14467 & 1.51158 & 1.17784 \\
            GES-GSP \textcolor{blue}{\fontsize{7}{10}\selectfont CVPR22}& 1.15462  & 1.11993 & \textbf{1.47197}  & 1.14340  & 1.04473 & 1.22256  \\
            \textbf{OBJ-GSP (ours)} & \textbf{0.98288} & \textbf{1.10622} & \textbf{1.08635} & \textbf{1.08296} & 1.47813 & \textbf{1.07699} \\
            \rowcolor{lightgray} Improvement(\%) & 5.9 & 9.5 & 5.9 & 3.3 & -0.4 & 5.8 \\
            \hline
            \multicolumn{7}{c}{Naturalness Image Quality Evaluator (NIQE $\downarrow$)} \\
            \hline
            UDIS \textcolor{blue}{\fontsize{7}{10}\selectfont TIP21}& 5.02442 & 3.76994 & 3.61888 & 4.67437 & 8.02090 & 4.35149 \\
            UDIS++ \textcolor{blue}{\fontsize{7}{10}\selectfont ICCV23}& 4.93279 & 3.66565 & 3.67661 & 4.38520 & 7.5419 & 4.1248 \\
            GSP \textcolor{blue}{\fontsize{7}{10}\selectfont ECCV16}& 3.84897 & 4.28546 & 3.18549  & 5.10784 & 7.00495 & 3.04781 \\
            GES-GSP \textcolor{blue}{\fontsize{7}{10}\selectfont CVPR22}& 3.79240 & 3.35315 & \textbf{2.75234} & 4.69390 & 6.96670 & 2.97173 \\
            \textbf{OBJ-GSP (ours)} & \textbf{3.70041} & \textbf{3.23057}  & 2.81480 & \textbf{4.08903} & \textbf{6.96149} & \textbf{2.49712} \\
            \rowcolor{lightgray} Improvement(\%) & 2.4 & 3.7  & -2.2 & 12.9 & 0.5 & 16.0 \\
\bottomrule
        \end{tabular}
    \end{center}
    \label{table_experiment}
\end{table*}

\section{Experiments}
\label{sec:experiments}


\subsection{StitchBench}
Previous work often collected a small number of images themselves and performed qualitative tests only.
Meanwhile, they have different focuses, such as parallax between the foreground and background, sparse features in natural scenery, precise alignment and no distinct structures to preserve, and distinct line structures, without comprehensively evaluating models' performance in a wide range of scenarios.
To address the issue, we present the most extensive image stitching benchmark to date: StitchBench, which include 122 pairs of images from 12 works. 
We collect 18 pairs of images captured by cameras, in which the preservation of object structures is crucial. StitchBench also includes 7 sets of urban scenes captured by low-altitude drones, featuring tall buildings and requiring the assistance of segmentation models. 
To overcome our subjective preferences and the limited locations where we collected the images, we also collect test images used in previous state-of-the-art works, namely AANAP~\cite{aanap}, APAP~\cite{apap}, CAVE~\cite{cave}, DFW~\cite{dfw}, DHW~\cite{dhw}, GES-GSP~\cite{gesgsp}, LPC~\cite{lpc}, SEAGULL~\cite{seagull}, REW~\cite{rew}, SVA~\cite{sva} and SPHP~\cite{sphp}. 
StitchBench is currently the most comprehensive stitching test dataset. An algorithm should demonstrate general applicability to perform well on all subsets of StitchBench: aligning well and preventing distortion naturally.


\textbf{Evaluation metrics}.
We quantitatively assess the quality of our stitching results from two perspectives: distortion prevention and alignment. 
First, we employ the Mean Distorted Residuals (MDR) metric to measure the degree of image distortion. 
In intuitive terms, if the points on the same side of the mesh were originally collinear and remain collinear after stitching, it implies that the stitching result has minimal distortion. 
Furthermore, we employ the Naturalness Image Quality Evaluator (NIQE)~\cite{niqe} metric to evaluate alignment performance. 
We argue that NIQE is a more intitutive and better indicator of alignment than RMSE, SSIM and PSNR, as it measures image clarity, and stitching results with misalignment will produce blurry areas, leading to worse NIQE scores.

\subsection{Baselines}
We compare with GSP~\cite{gsp} and GES-GSP~\cite{gesgsp}.
UDIS~\cite{udis} and UDIS++~\cite{udisplus} are famous works in applyig deep learning into image stitching. Since they do not explicitly use feature points, we are unable to measure its quality with MDR. 
We provide a detailed comparasion between OBJ-GSP and UDIS++ in the supplementary materials.

\subsection{Results}
\textbf{Quantitative results}.
Table \ref{table_experiment} shows MDR and NIQE results on datasets used in other stitching algorithms and our own dataset. We outperform GSP and GES-GSP in both alignment and shape preservation. UDIS++~\cite{udisplus} is a good try in deep learning based image stitching, but the performance is still no better than ours.

\textbf{Qualitative results}.

Fig.~\ref{fig_ges11} and~\ref{fig_ges2} elucidates the reasons behind the superior performance of our OBJ-GSP method. With the assistance of semantic segmentation techniques, we place greater emphasis on preserving critical structures and ensuring holistic protection at the object-level for objects within the images.
Fig.~\ref{fig_ges21} and~\ref{fig_rewgym} illustrates the stitching outcomes of six different methods, where we use straight lines and boxes to demonstrate the effects of alignment and distortion. 
\textbf{Please kindly refer to our supplementary material for more quanlitative results.}

\begin{figure*}[h]
	\begin{center}
    \includegraphics[width=1\linewidth]{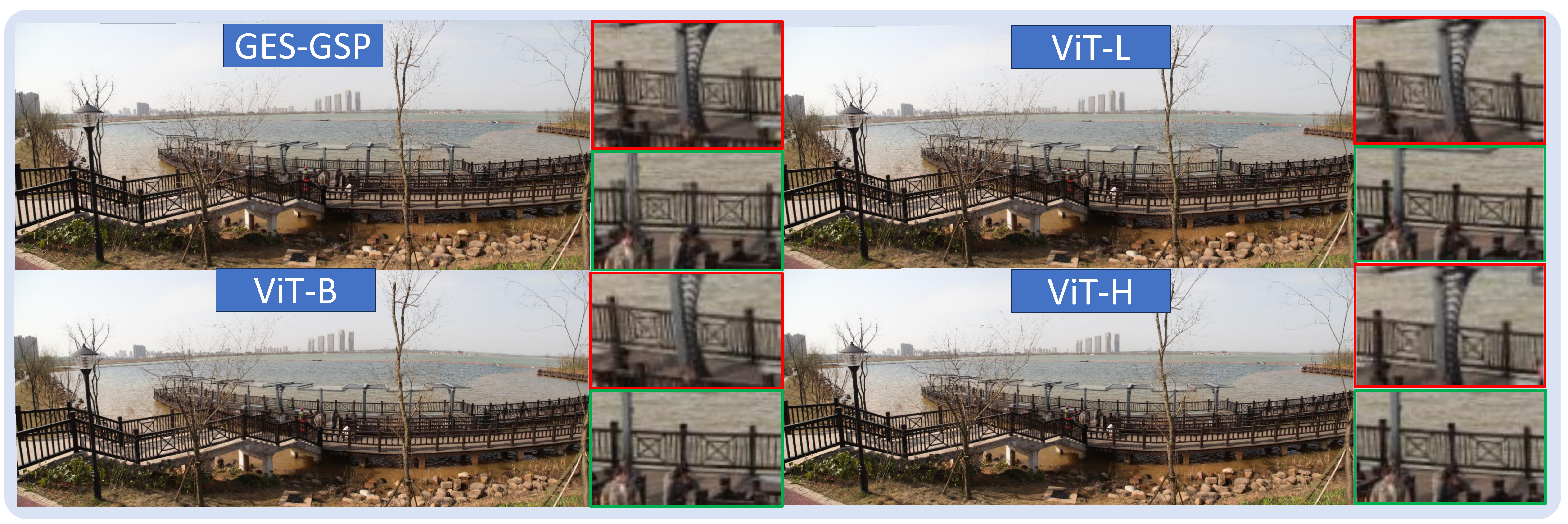}
	\end{center}
 	\caption{GES-GSP and the proposed OBJ-GSP with three Segment Anything Model backbones.}
	\label{fig_ablation}
\end{figure*}

\begin{figure}[h]
	\begin{center}
    \includegraphics[width=1\linewidth]{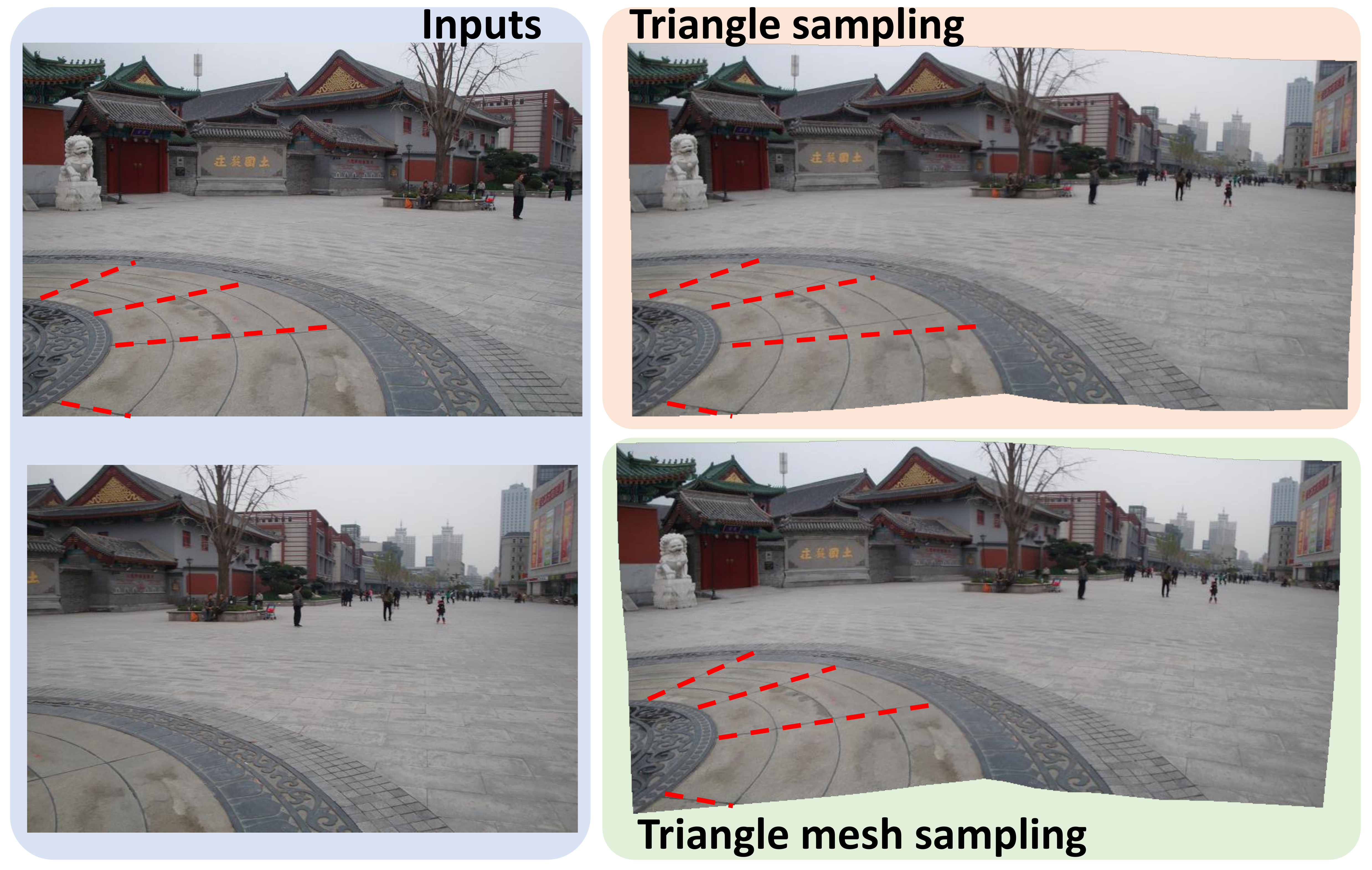}
	\end{center}
 	\caption{Triangle sampling preserves the shapes of lines on the ground, while triangle mesh sampling fails in preserving.}
	\label{fig_dhw2}
\end{figure}

\subsection{Low-Altitude Aerial Image Stitching}
Image stitching requires meeting one of two conditions~\cite{autostitch}: either the camera's optical center remains stationary while the camera rotates, or the scene only consists of objects that are far from the camera. 
Existing stitching algorithms mainly address the issue of stitching when these conditions are slightly violated. 
For low-altitude aerial images, where the flight height of the aircraft is around 100 meters but the height of buildings is no less than 20-40 meters, the camera's optical center moves significantly during drone shooting, thus completely failing to meet the two assumptions for stitching.
Moreover, if the left and right walls of a building are captured in two separate shots, it would be a logical error to include both walls in a panorama (for a cube, at most three faces can be seen at a time, and it's impossible to see two opposing faces simultaneously).
For stitching low-altitude aerial images, we first use a semantic segmentation model to segment the roofs and walls. We then calculate the height of the roofs and orthographically project the buildings onto the ground plane before stitching. 
In this scenario, the semantic segmentation model is essential for the stitching process~\cite{ISEO}.
The stitching pipeline and result is in Fig.~\ref{fig_aerialpipeline}.

\section{Ablation Studies and Discussions}
\label{sec:ablation}

\subsection{Lightweight SAM backbones}
To assess the influence of semantic segmentation results on the stitching performance, we conducted a comparative analysis across three different backbones of the SAM~\cite{sam} model, namely ViT-B, ViT-L, and ViT-H~\cite{vit}, with a progression from smaller to larger models. Larger models are inherently capable of capturing more fine-grained semantic details. The corresponding stitching results are depicted in Fig.~\ref{fig_ablation}. It is worth noting that the models based on ViT-B and ViT-L exhibit some blurriness and minor distortions. From the perspective of MDR, our model, under the three aforementioned backbone configurations, achieved improvements of 1.9\%, 2.7\%, and 3.6\% over the baseline model GES-GSP~\cite{gesgsp}, respectively. The application of EfficientSAM~\cite{efficientsam}, time consumption of OBJ-GSP, and its real-world applications are included in supplementary materials. 

\subsection{Sampling strategies} 
Incorporating SAM~\cite{sam} with ViT-H \cite{vit} backbone, triangular mesh sampling yields superior results compared to the triangular sampling proposed in GES-GSP~\cite{gesgsp}. 
The triangular mesh preserves the shape of lines but fails to maintain the overall geometric structure of the image. 
As illustrated in the Fig.~\ref{fig_dhw2}, we outline the structural elements in the original image using red dashed lines and then superimposed them onto the combined results of triangular sampling and triangular grid sampling. 
Triangular grid sampling retains the positional relationships between the lines present in the original image.

\section{Conclusion}
\label{sec:conclusion}

In this paper, we propose OBJect-level Geometric Structure Preserving for natural image stitching (OBJ-GSP) algorithm, which stands as a novel approach to achieving natural and visually pleasing composite images. 
OBJ-GSP protects object shapes by first segmenting them out, and then preserve the structures with triangle meshes.
We also demonstrate that semantic segmentation is necessary when it comes to low-altitude aerial image stitching.
We collect new test image pairs in common scenes and aerial imaging, and choose images from previous works, to establish the most comprehensive image stitching benchmark by far: StitchBench. Detailed experiments with comprehensive baselines in StitchBench demonstrate the effectiveness of OBJ-GSP.

\section{Extensive Discussions}
\label{sec:discussion}

\subsection{Limitations} 
While OBJ-GSP has demonstrated state-of-the-art performance in image stitching by extracting object-level geometric structures with semantic segmentation and preserving them with triangle mesh sampling, it introduces a large semantic segmentation model into image stitching, resulting in higher computational costs. 
However, with the development of semantic segmentation techniques, lighter versions of SAM will emerge~\cite{efficientsam,lightweightsam}, enhancing the speed of our work.
According to our analysis, the effectiveness of geometric structure extraction significantly impacts the final results. Our method is constrained by the quality of SAM's results. Smaller models like SAM~\cite{sam} with Vit-B/L~\cite{vit} do not perform as SAM with Vit-H. 
To stitch a pair of 800*600 images, SAM spends 25s on RTX2090 with 8-24G GPU memory, depending on the backbone. For C++ ONNX implementation, SAM VIT-B spends 1.5 min on CPU.
Mesh optimization and image processing cost less than 4s on an Intel i5 CPU, almost the same as GES-GSP~\cite{gesgsp}.
OBJ-GSP needs more computational resources and time than GES-GSP, but the stitching quality is also better. 
The time cost of SAM is larger than line detection methods in GES-GSP. 
The time used for triangle mesh optimization is almost the same as that in GES-GSP.

\subsection{Applications of OBJ-GSP}
In many fields, there is a need for high-quality stitched images, even at the expense of long time costs and significant computational resources.
In medical image processing~\cite{medical1,medical2,medical3}, for instance, stitching multiple pathological slice images to reconstruct the entire tissue structure or organ's three-dimensional model demands high precision and quality. These tasks typically necessitate precise alignment and seamless fusion, and can tolerate longer computational times to ensure accuracy. Similarly, in photography~\cite{photography} or virtual tourism~\cite{vt1,vt2} applications, stitching numerous high-resolution images is necessary to generate high-quality panoramic images. This holds true in the fields of remote sensing~\cite{rs1,rs2,rs3} and movie industry~\cite{movie1,movie2} as well.
Currently, the emergence of faster and more accurate segmentation models, such as EfficientSAM~\cite{efficientsam}, makes our method even more promising. We provide detailed comparisons in the supplementary materials.

\subsection{Post-processing in image stitching} 
Our method primarily addresses computing transformation matrices to achieve alignment and shape preservation. Post-processing techniques can be combined with our approach to achieve a more natural stitching effect. Blending~\cite{autostitch,blend1,blend2} and seam-driven~\cite{seam} methods can be used to further reduce blurring, while global straightening~\cite{dhw} can decrease distortion.

\subsection{Broader impacts} It is important to acknowledge that we do not explicitly discuss broader impacts in the proposed OBJ-GSP image stitching algorithm, such as fairness or bias. Segment Anything Model (SAM) \cite{sam} has discussed its broader impacts regarding geographic and income representation, as well as Fairness in segmenting people. Further research into how our algorithm may interact with other aspects of image stitching is encouraged.

\section{Ablation of EfficientSAM and mesh sampling}
\begin{figure*}[h!]
	\begin{center}
    \includegraphics[width=1\linewidth]{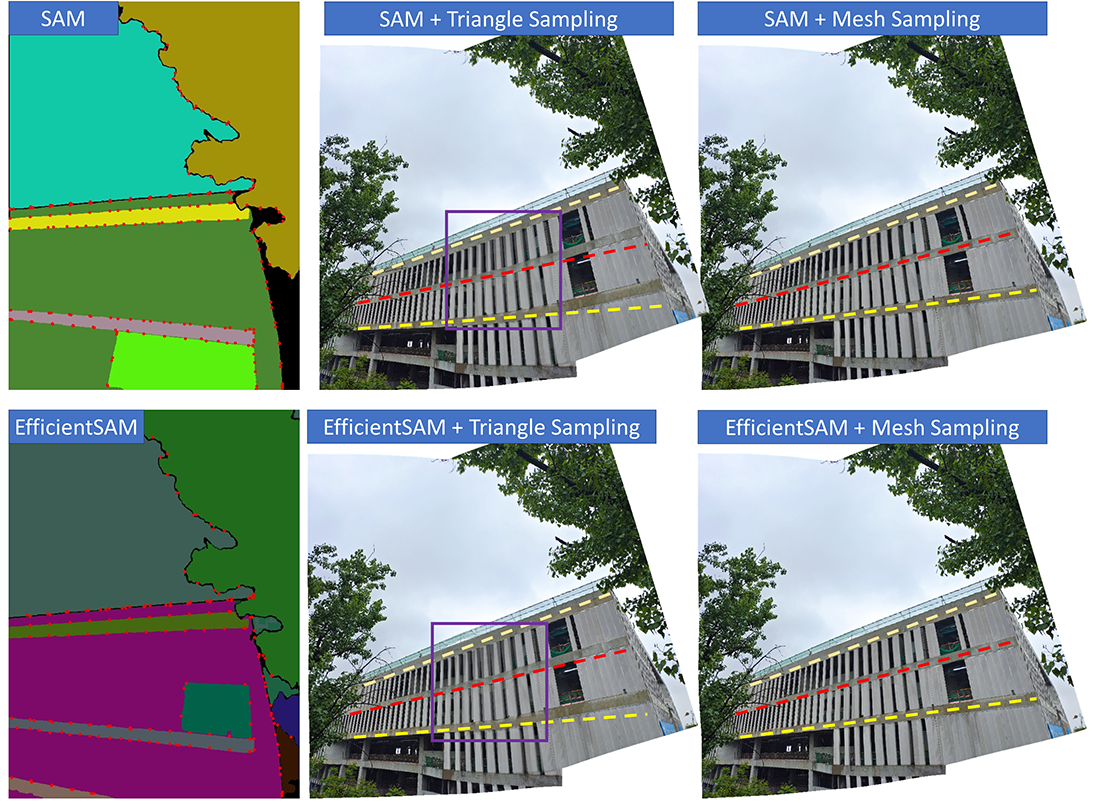}
	\end{center}
 	\caption{\textbf{Ablation study of EfficientSAM~\cite{efficientsam} and mesh sampling.}. The left images shows SAM~\cite{sam} and EfficientSAM~\cite{efficientsam} segmentation results and sampling points on them. The purple boxes indicate misalignment in triangle sampling methods. Also, both triangle mesh based methods undergo distortion. We see no apparent difference between SAM~\cite{sam}+mesh sampling and EfficientSAM+mesh sampling results.}
	\label{fig_efficientsam}
\end{figure*}

We propose the utilization of SAM-based methods and mesh sampling to address distortion and misalignment during stitching. It is important to emphasize that both components are indispensable for object-level shape preservation. Fig.~\ref{fig_efficientsam} illustrates the distinct results achieved under the same semantic segmentation output using line-based triangle sampling and object-level mesh sampling. Mesh sampling can recognize object structures and effectively preserve objects from distortion.
Furthermore, with the advancement in the field of semantic segmentation, the speed of SAM-based methods is accelerating, which will greatly expedite our image stitching approach. For instance, EfficientSAM~\cite{efficientsam}. In Fig.~\ref{fig_efficientsam}, there is no significant difference observed between the results obtained using SAM + mesh sampling and EfficientSAM + mesh sampling. However, the time consumption of EfficientSAM is only $5\%$ of SAM, which is predictable. With the further development of SAM-based methods, even faster and more accurate approaches are expected to emerge, making our stitching method faster and more precise.

\section{StitchBench Metadata}
\label{sec:dataset}
We employed handheld smartphones to capture OBJ-GSP images. During the image acquisition process, we took care to minimize translational movement of the smartphone, primarily relying on rotation to adjust the framing. 
This approach was employed to ensure that the disparity between images remained manageable, preventing situations where the occlusion relationships between two images would be too dissimilar for successful image stitching. 
We amassed a total of 18 pairs of image sets, encompassing diverse scenes such as rooms, culinary creations, sculptures, gardens, rivers, ponds, industrial facilities, roads, and exteriors of buildings, among others. 

We also collect 7 sets of aerial images, each consist of about 9 pairs of images of urban scenes. We fly drones at 100-120 meters, in urban areas where there are buildings, roads, trees, etc~\cite{ISEO}.
We collect images with a DJI Mavic Air 2 and the image size is set to be 3000 × 4000 Pixels. Cameras on the drone are kept vertical to the ground (bird view).

Additionally, we curate existing image stitching datasets \cite{aanap, apap,cave,dfw,dhw,gesgsp,lpc,seagull,rew,sva,sphp}
to supplement our data collection efforts. 
Ultimately, we constructed a dataset consisting of 122 groups of images, marking it as the largest dataset currently employed in image stitching endeavors. 
We release this dataset to the public for further research and development.

\section{Reestablishing baselines}
\label{sec:reestablish}

\subsection{Implementation details} We evaluated the results of GSP \cite{gsp} and GES-GSP \cite{gesgsp} using their publicly available C++ code. Our stitching code is also implemented in C++. Our implementation of SAM includes two approaches. Drawing from the Segment Anything C++ Wrapper \cite{cppsam}, we exported the SAM \cite{sam} encoder and decoder into Open Neural Network Exchange (ONNX) format and subsequently replicated SAM's automatic mode within C++. To achieve the best possible stitching results, we also directly implemented the semantic segmentation component using the SAM's publicly available Python code and utilized their semantic segmentation results. The stitching part of our experiments ran on the CPU, while the SAM modules were capable of running on both CPU and GPU. 

\subsection{Baselines}
We replicated the results reported in the literature for GSP \cite{gsp}, GES-GSP \cite{gesgsp}, APAP \cite{apap}, and SPHP \cite{sphp} by implementing their publicly available codebases with their default parameter settings. For the structural alignment component, we employed the executable provided by Autostitch \cite{autostitch}. GES-GSP includes experimental data for both GES-GSP and GSP. 
In our method's experiments, we maintained consistent parameter settings across all trials. Furthermore, for the structure extraction stage, we utilized the official code provided by SAM. However, we excluded masks with extremely small areas.

\section{Algorithm and Results of Low-Altitude Aerial Image Stitching}
\subsection{Why is Semantic Segmentation Necessary}
\begin{figure}[h!]
	\begin{center}
    \includegraphics[width=1\linewidth]{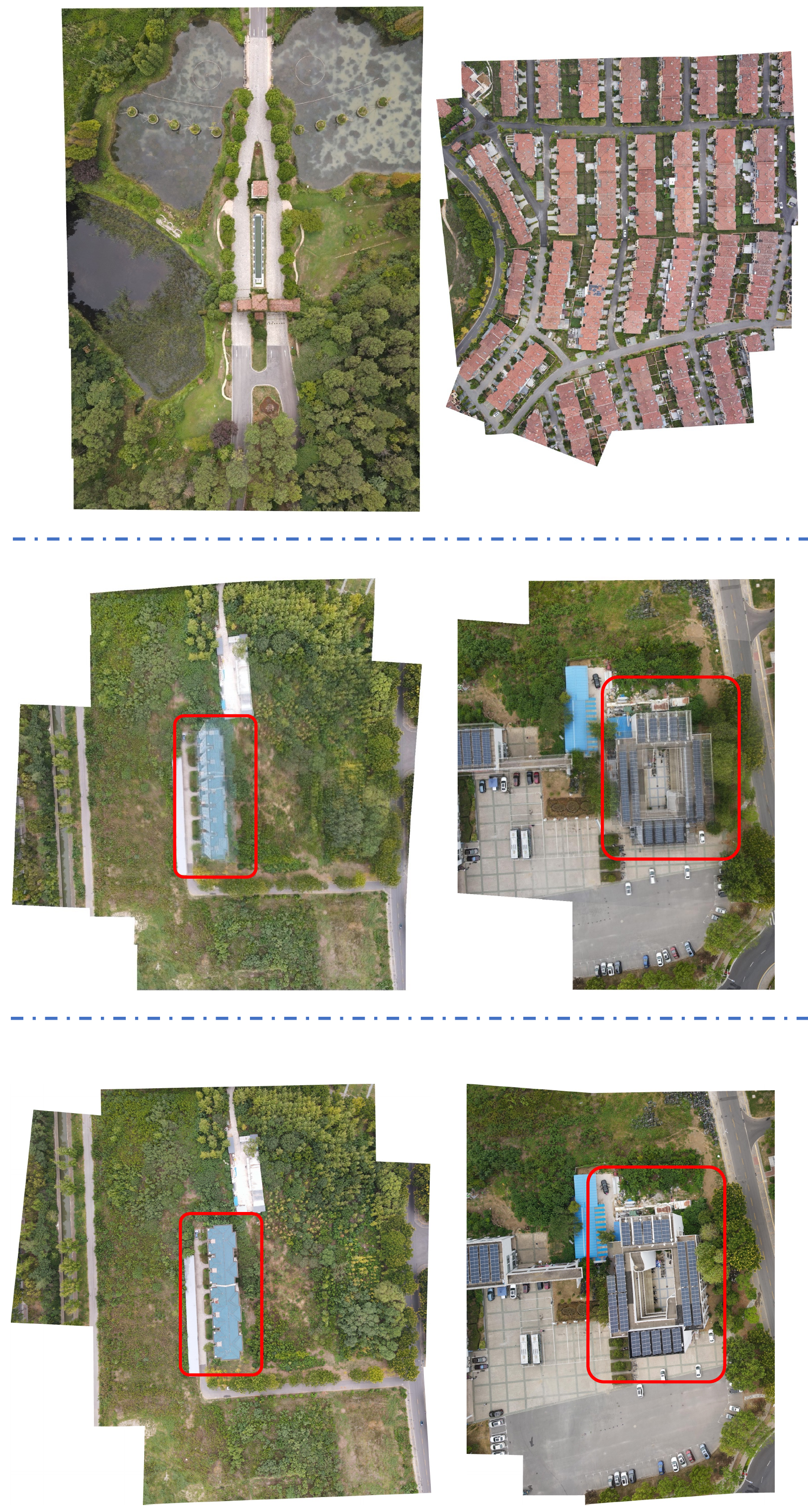}
	\end{center}
 	\caption{In these images, drones fly at about 100m height. \textbf{Top}: segmentation and orthographic projection is not necessary when everything in the images are relative low. In top left image, the captures scene is almost a plain, and the towers is about 6 meters tall. The buildings are two-floor villas in the bottom left image, and no more than 8 meters tall. \textbf{Middle}: The building in the middle left images are 6 floors, and about 20 meters tall. In middle right images, the buildings are about 15 meters tall. The tall buildings can not be stitched by traditional stitching methods. \textbf{Bottom}: OBJ-GSP can stitch the images by segmenting roofs and walls out, transform roofs with orthographic projection,and then stitch.}
	\label{fig_noortho}
\end{figure}

For low-altitude drone aerial photography in urban areas, the drone's flight altitude is low while the buildings are tall. Due to the significant distance difference between the drone camera and the buildings, the transformation matrices for buildings and rooftops differ from those for the ground in different images. Direct stitching can lead to a lot of ghosting and unnatural, distorted building structures. Additionally, selective information must be discarded during low-altitude aerial stitching in urban areas. If buildings are simply considered rectangular prisms, and the left and right walls are captured in two separate shots, it is impossible to retain both sides in the stitched image (as a person cannot see both opposing sides of a rectangular prism simultaneously). Therefore, this paper proposes first using a segmentation model to identify buildings and walls in the scene. The walls are removed from the images, and the ratio of the building height to the drone height is calculated based on the different transformation relationships for the ground and buildings. The buildings are then projected onto the ground plane before stitching.
The importance of segmentation in aerial image stitching is shown in Fig.~\ref{fig_noortho}


\subsection{OBJ-GSP in Aerial Image Stitching}
In aerial images of low altitude, there are two planes of interest: ground $P^g$ and roofs $P^r_i$, where $i=1,2,...N$ and $N$ is the number of roofs. Semantic segmentation models are adopted to segment roofs and walls, with the remaining pixels regarded as ground. We mask out walls, and then orthographically project roofs to grounds. 
For correctly matched feature points ${f}_{r1}$ and ${f}_{r2}$on $P^r$, and ${f}_{g1}$ and ${f}_{g2}$ on $P^g$, we aim to find transformation matrix $H_o$ to project roof to ground before stitching. After projection, the images can be stitched with a global transformation matrix, which is also the transformation between grounds, $H_g$.

\begin{equation}
\label{eq:1}
	\begin{cases}
		f_{g2}=H_{g}f_{g1} \\
		H_{oi}f_{r2}={H_g}H_{oi}f_{r1}  
	\end{cases}
 \text{for}\ i \ \text{in} 1\ \text{to}\ N\text{,}
\end{equation}

where $f$ is in homogeneous coordinates $ \left[\begin{matrix}x&y&1\\\end{matrix}\right]^T$. 
Let the height of roof $P^r_i$ to the ground be $h_{ri}$, and height of drone be $h_d$, for each pixel $[x,y,1]$ on the roof, the orthographic projection transformation is 

\begin{equation}
		\left[\begin{matrix}x_g\\y_g\\1\\\end{matrix}\right]=\ \left[\begin{matrix}1-\frac{h_{ri}}{h_d}&0&\frac{h_{ri}}{h_d}a\\0&1-\frac{h_{ri}}{h_d}&\frac{h_{ri}}{h_d}b\\0&0&1\\\end{matrix}\right]\ \left[\begin{matrix}x_r\\y_r\\1\\\end{matrix}\right] 
 \text{for}\ i \ \text{in} 1\ \text{to}\ N\text{,}
\end{equation}
where $a$ and $b$ are half of pixel width and height of the image.
If the transformation matrix is in the form of a homograph matrix, and the number of feature point matches is $M_i$ on $P^r_i$, Eq.~\ref{eq:1} expands to $2M_i$ mutually independent quadratic equations with the unknown $\frac{h_{ri}}{h_d}$. This over determined equation can be solved by methods such as Newton iteration. After solving  $\frac{h_{ri}}{h_d}$, the orthographic projection map can be generated by simply transforming $P^r_i$ to the ground one by one. The transformation matrix is the affine transformation matrix, which can be solved similarly for the orthographic projection matrix.

\subsection{Segmentation Models and Aerial Segmentation Datasets Used}
We finetune Grounded SAM~\cite{groundedsam} on low-altitude drone datasets where roof and wall are annotated (Varied Drone Dataset~\cite{VDD} and ICG Drone Dataset~\cite{icg}).


\section{More qualitative results}
\label{sec:moreresult}
\begin{figure*}[h]
	\begin{center}
    \includegraphics[width=1\linewidth]{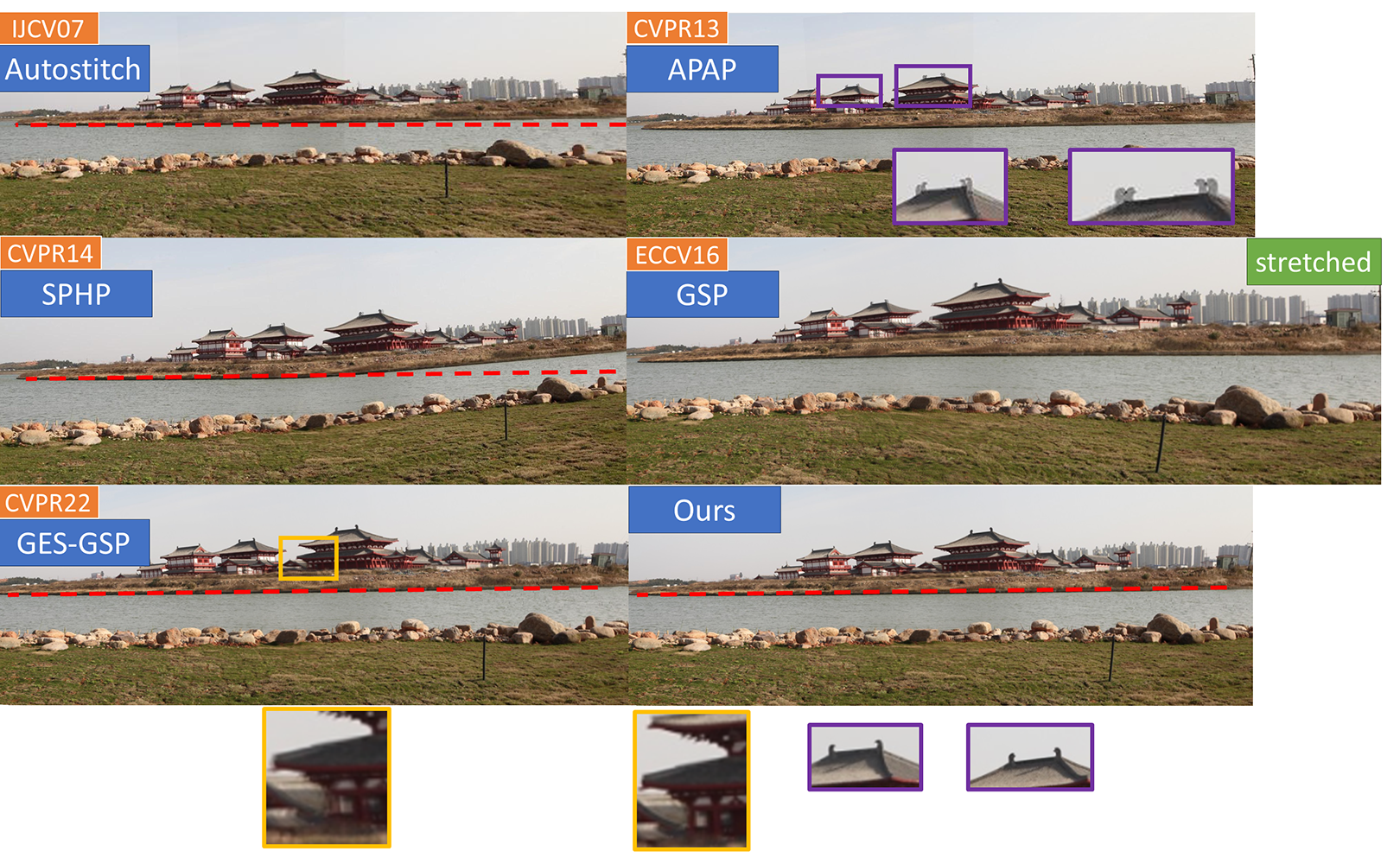}
	\end{center}
 	\caption{\textbf{More qualitative results}. Images are from LPC \cite{lpc} dataset. As marked with the red line,  Autostitch, SPHP and GES-GSP are distorted during transformation. APAP and GES-GSP sees misalignment. OBJ-GSP protects the shape of river bank and precisely aligns input images.}
	\label{fig_LPC15}
\end{figure*}

\begin{figure*}[h]
	\begin{center}
    \includegraphics[width=1\linewidth]{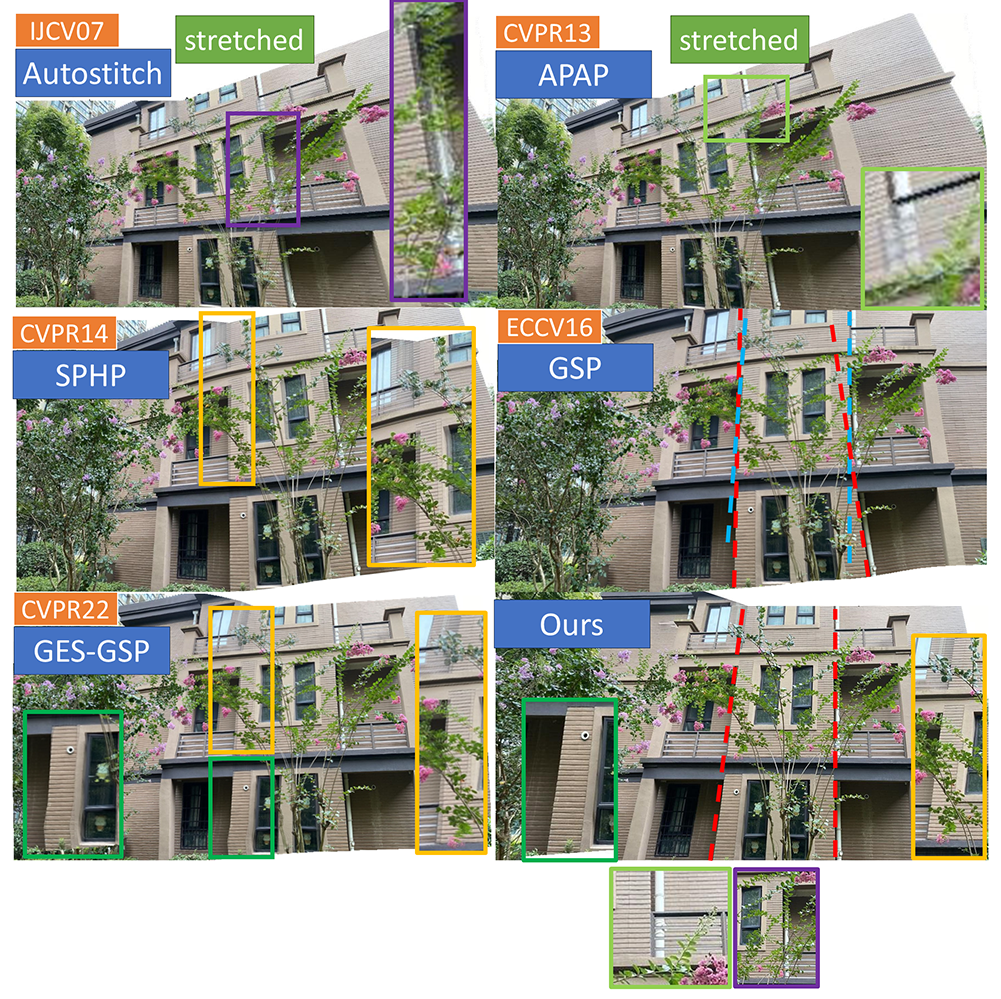}
	\end{center}
 	\caption{\textbf{More qualitative results}. Images are from LPC \cite{lpc} dataset. Almost all methods except from GSP and OBJ-GSP misalign images. GSP distorts the shape of building during transformation. OBJ-GSP tries to reach a balance between shape preservation and alignment, although there's some slight distortion in the green box.}
	\label{fig_lpc23}
\end{figure*}

\begin{figure*}[h]
	\begin{center}
    \includegraphics[width=1\linewidth]{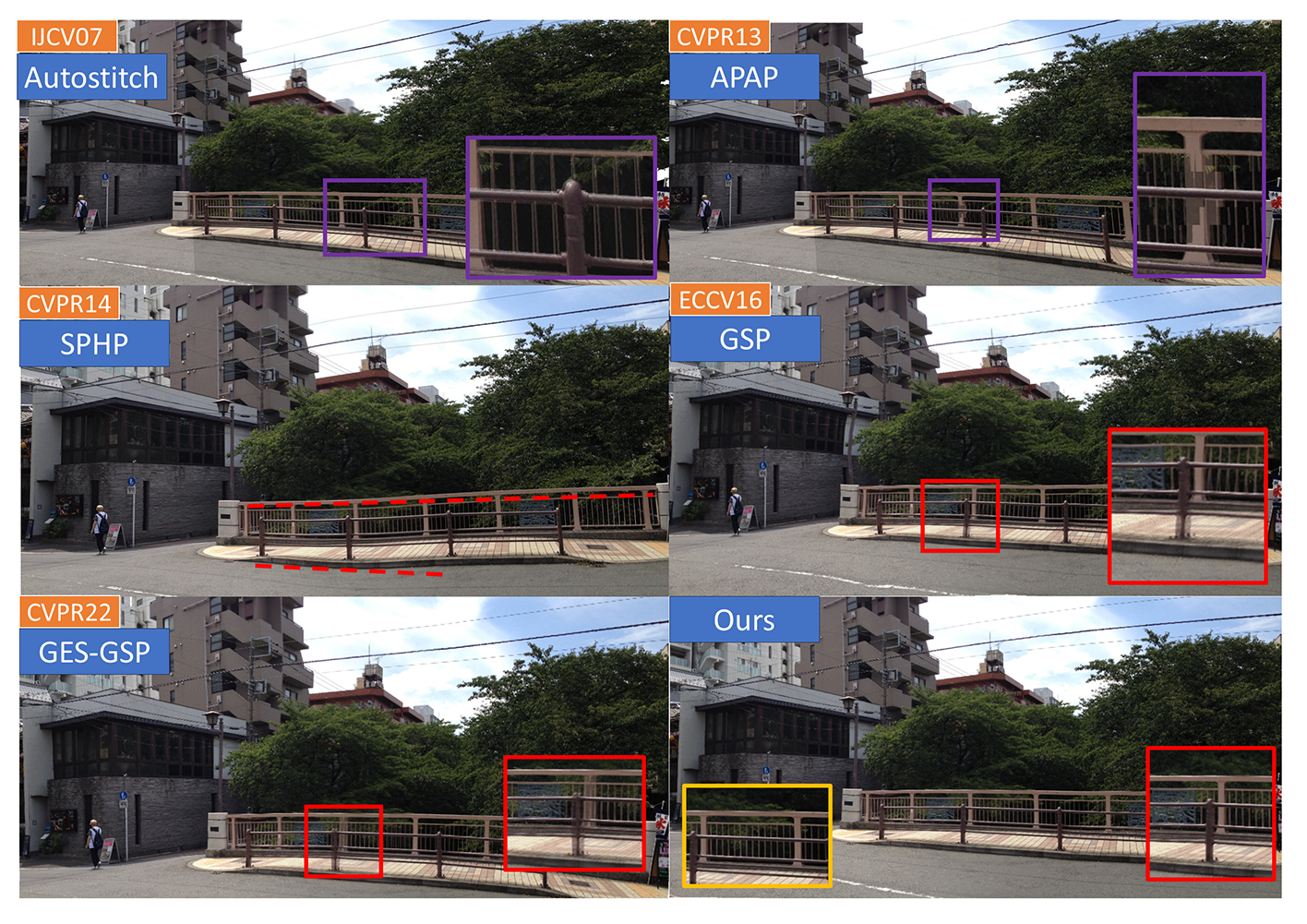}
	\end{center}
 	\caption{\textbf{More qualitative results}. Images are from SPHP \cite{sphp} dataset. GSP and GES-GSP suffers from misalignment as shown in the red boxes. The misalignment of Autostitch and APAP are shown in purple boxes. Also, SPHP distorts the bridge and ground.}
	\label{fig_sphp4}
\end{figure*}

\begin{figure*}[h]
	\begin{center}
    \includegraphics[width=0.8\linewidth]{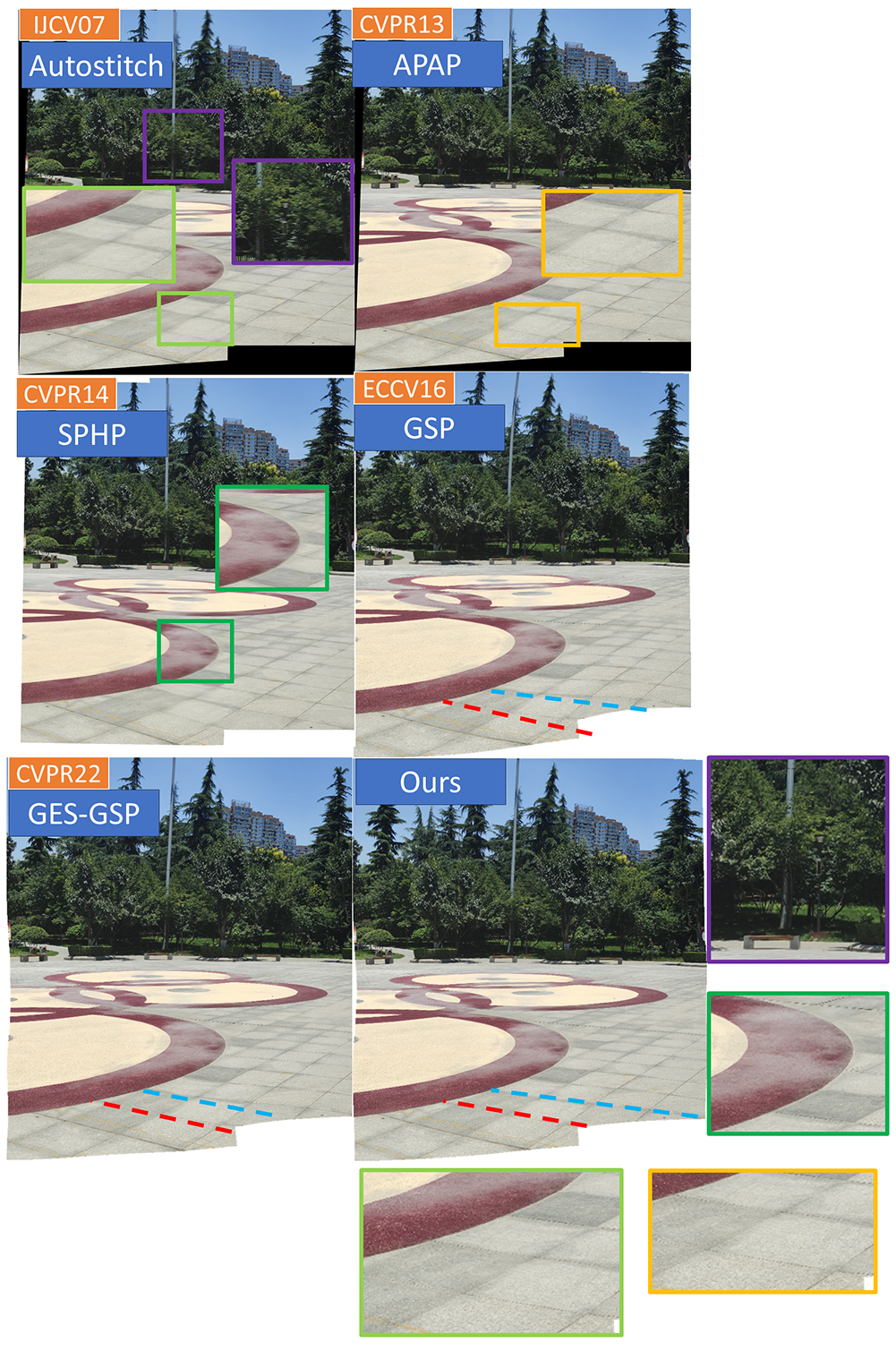}
	\end{center}
 	\caption{\textbf{More qualitative results}. Images are from GES-GSP \cite{gesgsp} dataset. Misalignment can be seen in results of Autostitch, APAP and SPHP. GES-GSP also distorts images. The stitched result of GSP is stretched. In contrast, the proposed OBJ-GSP aligns images well and protects pbject structures.}
	\label{fig_ges16}
\end{figure*}

In this section we provide more qualivative results. We mark images with \textbf{boxes} to indicate \textbf{misalignment}, and use \textbf{lines} (and intersections of lines) to show \textbf{distortion}. Please refer to Fig. \ref{fig_LPC15}, \ref{fig_lpc23}, \ref{fig_rewgym}, \ref{fig_sphp4} and \ref{fig_ges16}. It is shown that our object-level preservation of structures can prevent distortion and misalignment at the same time.

\section{Successful cases because of Segmentation}
\label{sec:sam-success}
\begin{figure*}[h]
	\begin{center}
    \includegraphics[width=1\linewidth]{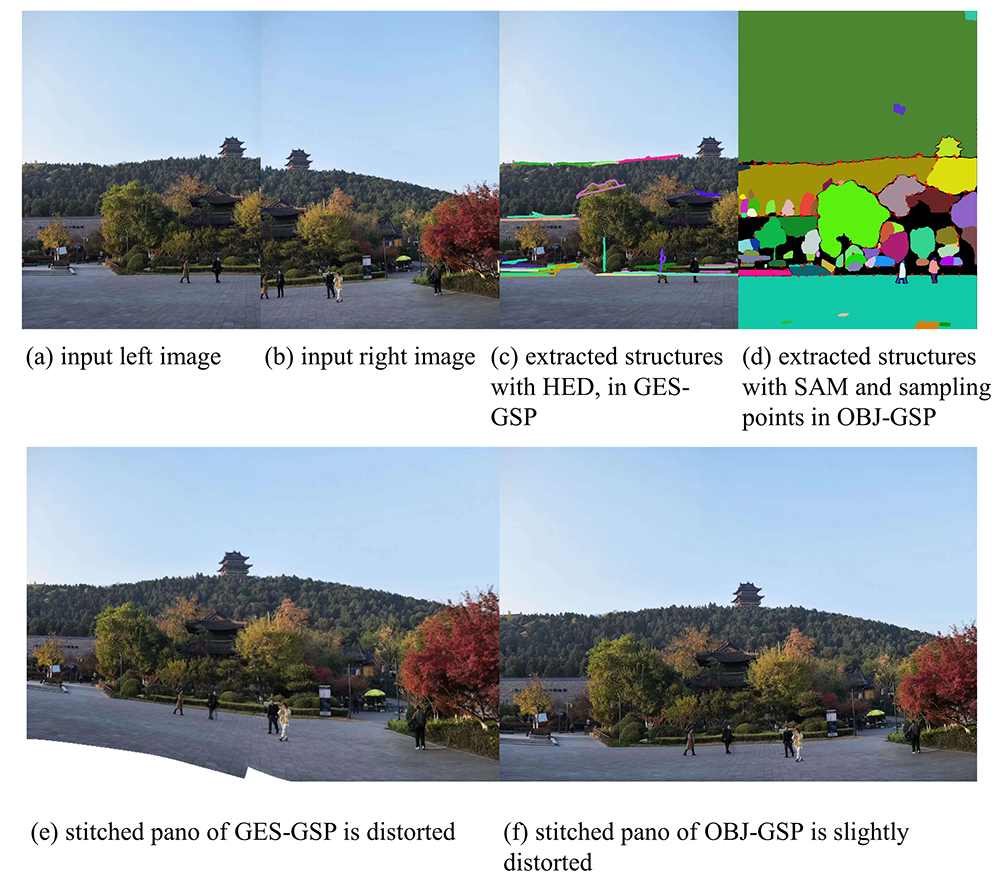}
	\end{center}
 	\caption{\textbf{SAM successful case}. Compared to HED \cite{hed}, SAM extracted complete structures of the mountain and ground. As a result, OBJ-GSP can protect their structures better, although the left part of the ground is slightly distorted.}
	\label{fig_temple}
\end{figure*}

\begin{figure*}[h]
	\begin{center}
    \includegraphics[width=1\linewidth]{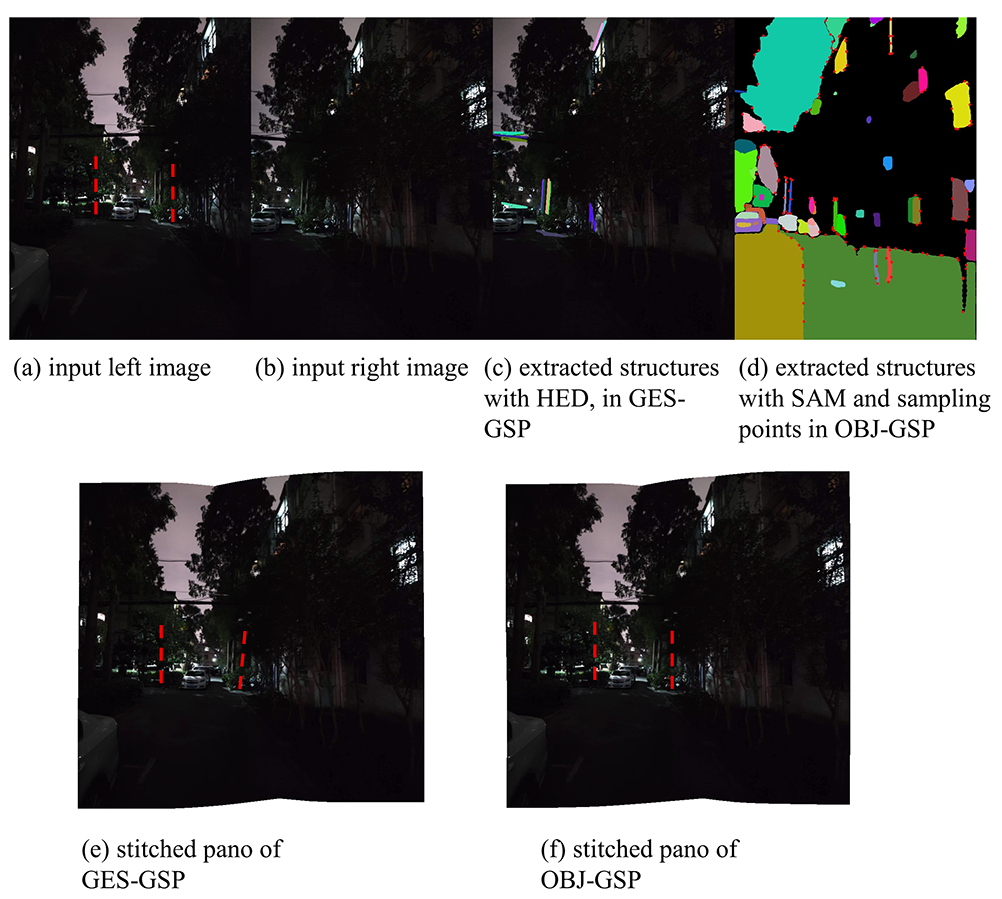}
	\end{center}
 	\caption{\textbf{SAM successful case}. Thanks to structures extracted by SAM, OBJ-GSP is able to stitch images taken at nights, with limited features. OBJ-GSP protects the structure of trees, as marked with red lines, because SAM's structures are more informative than HED's \cite{hed}.}
	\label{fig_night3}
\end{figure*}

In this section, we demonstrate the superiority of Segment Anything Model \cite{sam} with qualitative results. Please refer to Fig. \ref{fig_temple}  It is shown that SAM extracted object-level and complete structures of the ground and  mountain, so the OBJ-GSP preserved their structures better than GES-GSP, where HED \cite{hed} only extracted fragmented edge information. 
In Fig. \ref{fig_night3}, we demonstrate that the proposed OBJ-GSP can even stitch images with poor lighting conditions and protect their object-level structures.

\section{Failure cases}
\label{sec:sam-fail}

\begin{figure*}[h]
	\begin{center}
    \includegraphics[width=0.8\linewidth]{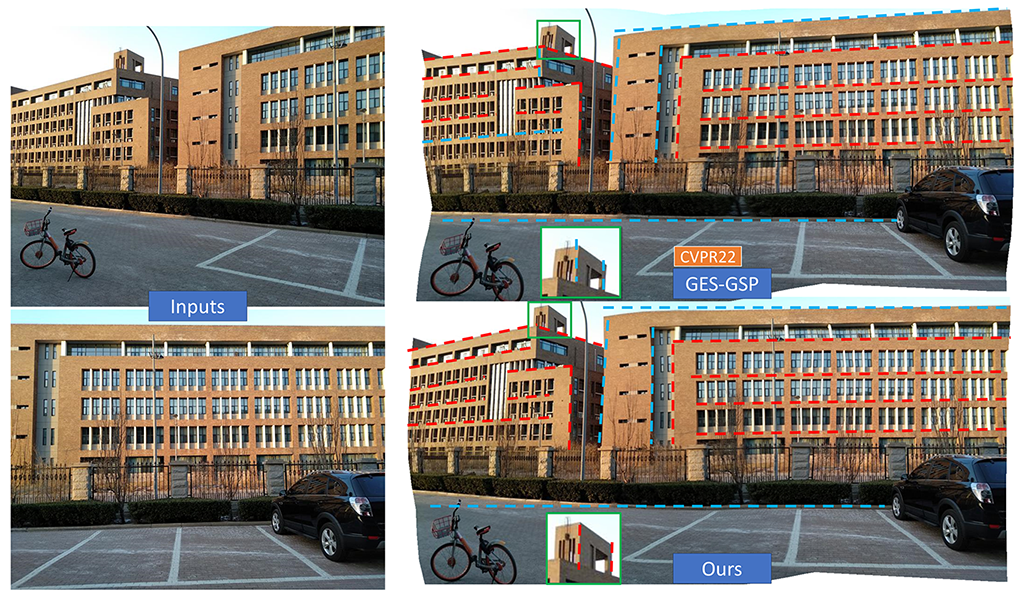}
	\end{center}
 	\caption{Failure case. Preserved structures are marked with red lines, while structures marked with blue lines are distorted due to considerable disparity in their perspectives within the input images. Although OBJ-GSP successfully preserve structures in green box, we see no apparent overall improvement over GES-GSP. }
	\label{fig_fail1}
\end{figure*}

As shown in Fig. \ref{fig_fail1}, OBJ-GSP fails in cases of significant parallax. In the perspective of the left image, the corner of the building appears to be obtuse. However, based on the inference from the right image, the corner of the building should be a right angle. Consequently, the stitching algorithm becomes perplexed, unsure of how to preserve the shape of the house. 

\begin{figure*}[h]
	\begin{center}
    \includegraphics[width=0.8\linewidth]{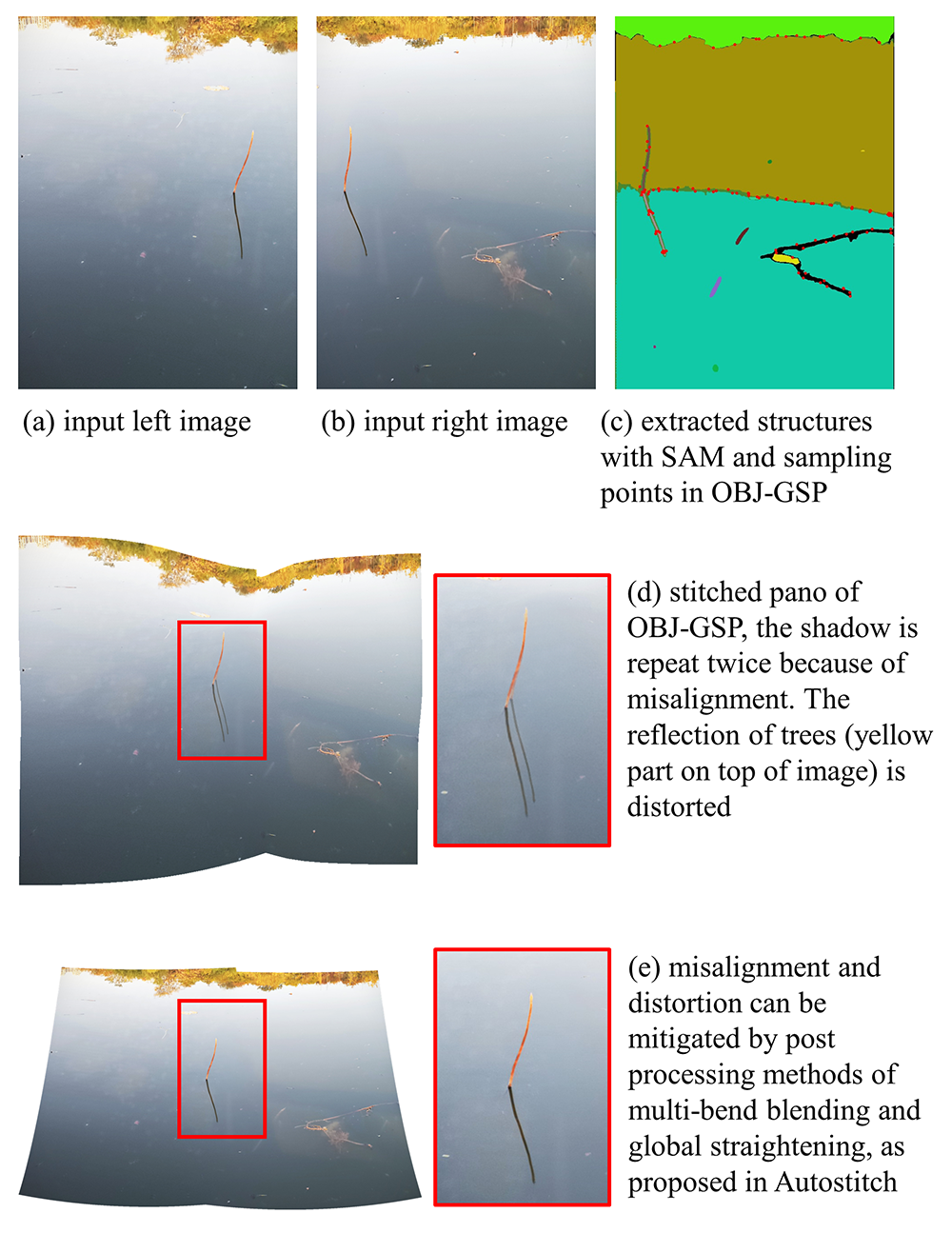}
	\end{center}
 	\caption{\textbf{SAM failure case}. The stitched pano of OBJ-GSP is distorted and misaligned because of the failure of SAM, but it can be metigated by global straightening and multi-bend blending.}
	\label{fig_water}
\end{figure*}

\begin{figure*}[h]
	\begin{center}
    \includegraphics[width=0.8\linewidth]{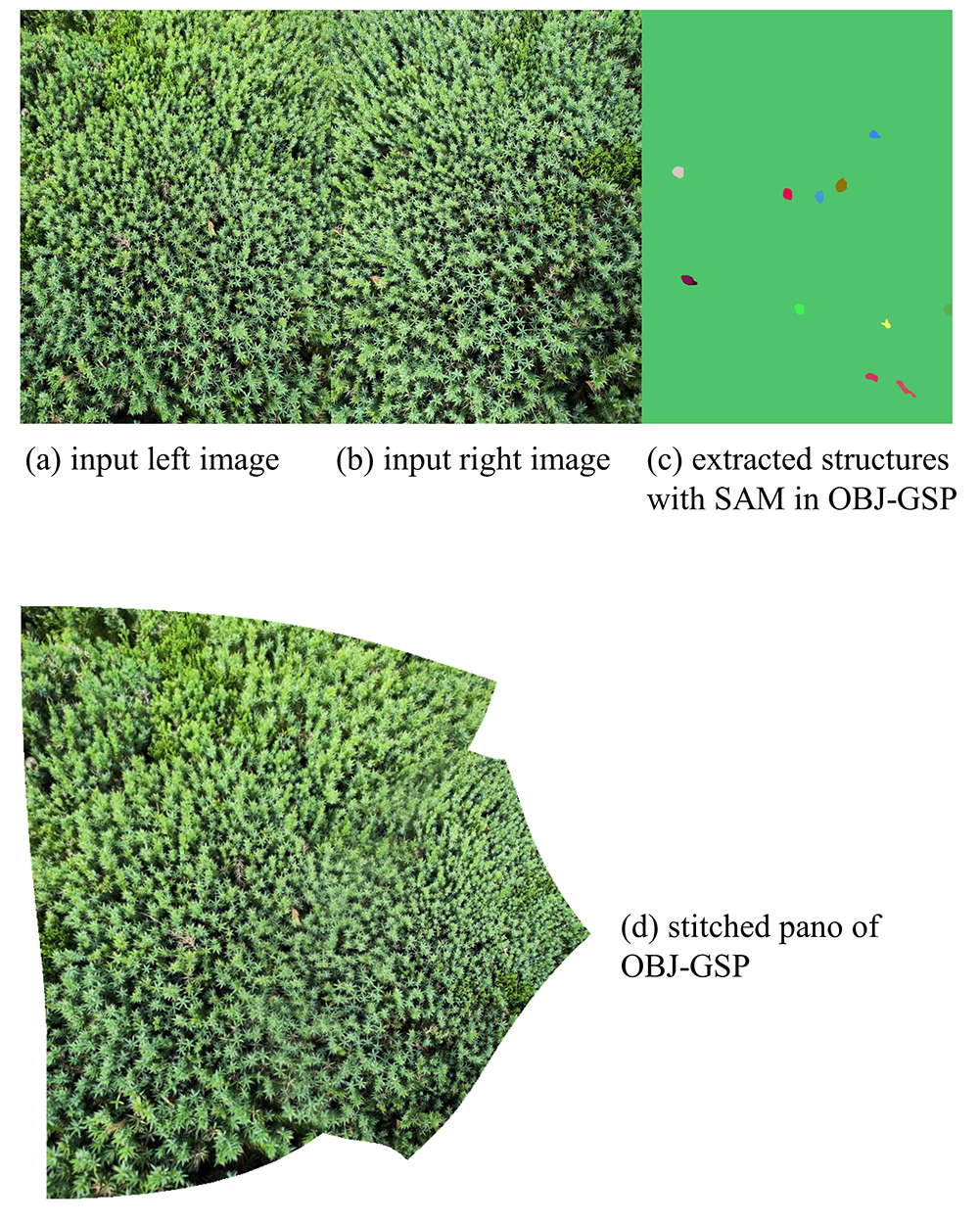}
	\end{center}
 	\caption{\textbf{SAM failure case}. SAM fails to extract meaningful object-level structures, so OBJ-GSP performs at the same level of GSP \cite{gsp}. }
	\label{fig_tree2}
\end{figure*}

In only rare instances OBJ-GSP experiences failures due to the malfunction of SAM \cite{sam}, such as in conditions characterized by inadequate illumination or an exceedingly sparse set of features. 
However, we acknowledge that it's very important to discuss the situations where, Segment Anything Model \cite{sam} fails, leading to the failure of the proposed OBJ-GSP image stitching algorithm. Please refer to Fig. \ref{fig_water} and \ref{fig_tree2}. 
In the context of our approach, it is not imperative for the SAM \cite{sam} to achieve precise segmentation of objects containing semantic information. SAM need only recognize key object contours and boundaries. Therefore, the proposed OBJ-GSP is susceptible to SAM failure only in exceedingly rare instances where features are exceptionally sparse, and objects are highly indistinct. Consequently, SAM's failure would impact solely the maintenance term of our structure, leading OBJ-GSP to degrade to the performance level of the GSP \cite{gsp}.
Simultaneously, we emphasize that SAM's failure would result in the nullification of our structural preservation term, causing OBJ-GSP's performance to regress to that of GSP only under scenarios where SAM proves ineffective. In addressing cases involving distortion and misalignment, we posit that mitigation strategies such as global straightening and multi-bend blending, as employed in Autostitch \cite{autostitch}, can be leveraged to alleviate these issues.

\section{Comparasion with UDIS++}
\begin{figure*}[h]
	\begin{center}
    \includegraphics[width=1\linewidth]{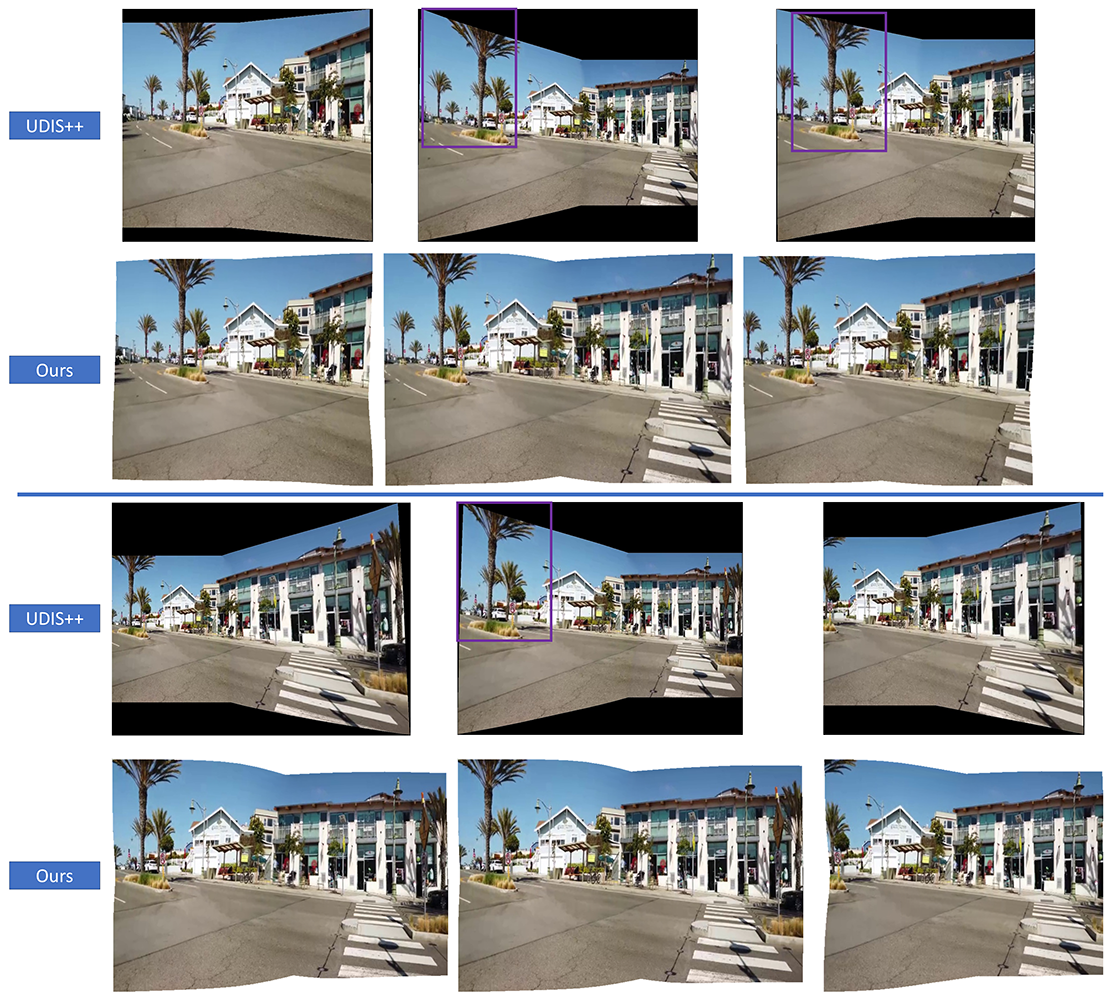}
	\end{center}
 	\caption{\textbf{Comparasion with UDIS++} on 6 sets of input images of a similar scene in its dataset.  Line 1 and 3 are UDIS++ results, while line 2 and 4 are ours. Purple box indicates distortion of tree.}
	\label{fig_udis1}
\end{figure*}

\begin{figure*}[h]
	\begin{center}
    \includegraphics[width=1\linewidth]{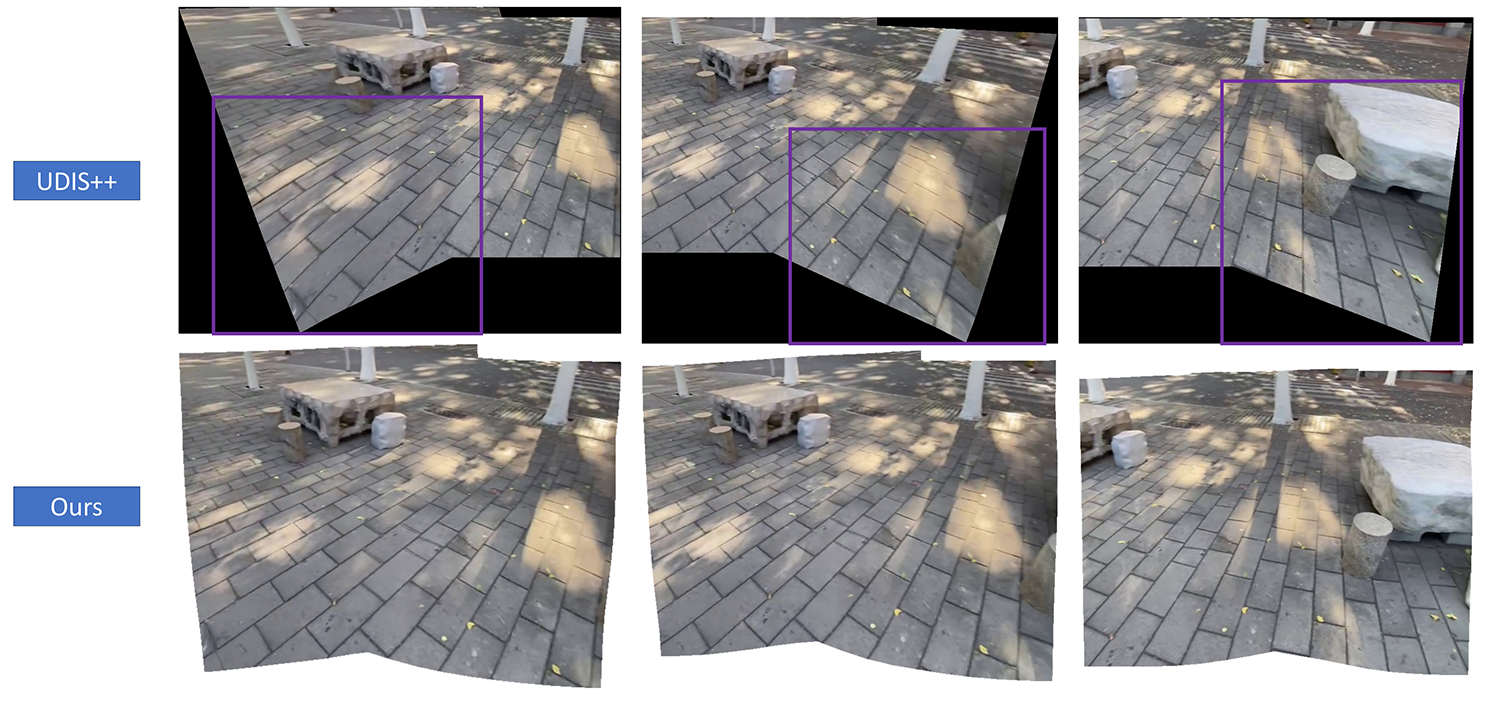}
	\end{center}
 	\caption{\textbf{Comparasion with UDIS++} in its dataset. Purple box indicates distortion: the floor tiles are stretched.}
	\label{fig_udis2}
\end{figure*}

\begin{figure*}[h]
	\begin{center}
    \includegraphics[width=1\linewidth]{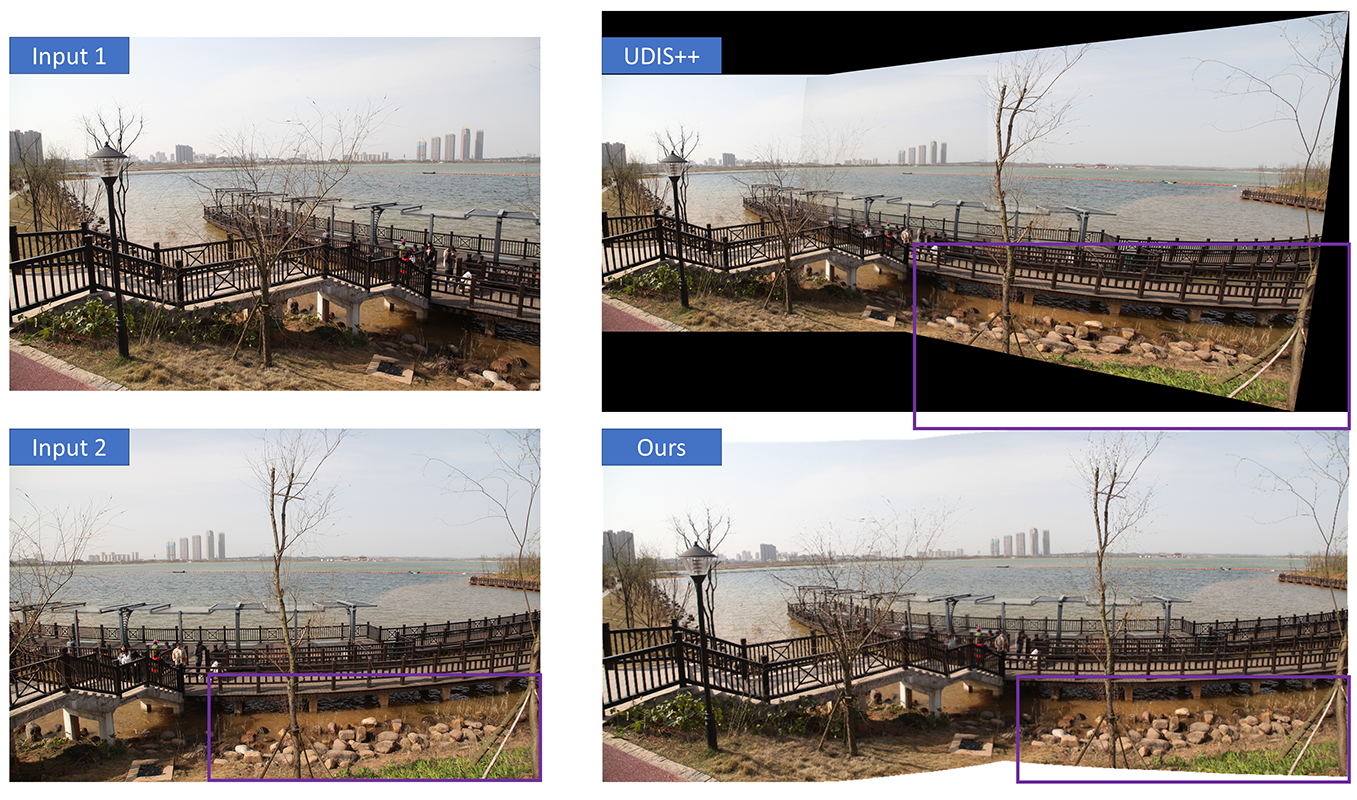}
	\end{center}
 	\caption{\textbf{Comparasion with UDIS++} in REW dataset~\cite{rew}. UDIS++ stretches the stones and ground in purple box, while OBJ-GSP preserves the global structure. }
	\label{fig_udis3}
\end{figure*}

UDIS~\cite{udis}(TIP 2021) and UDIS++~\cite{udisplus}(ICCV 2023) are a family of attempts to address image stitching problems using deep learning frameworks. 
We compare with the revised version (UDIS++) as it performs better than UDIS.
Like us, UDIS++ also aims to solve distortion issues on top of alignment. In the main text, we compute UDIS++'s NIQE to assess its alignment performance. Since UDIS++ is not feature point based, it cannot calculate MDR, a metric for measuring distortion. Therefore, in the supplementary material, we provide results for four scenes with multiple sets of images to intuitively compare distortion levels. Our method outperforms in both distortion resilience and alignment. Please refer to Fig. \ref{fig_udis1},  \ref{fig_udis2} and  \ref{fig_udis3}.

\clearpage 

\clearpage 

\clearpage 

\pagebreak

{
	\small
	\bibliographystyle{ieeenat_fullname}
	\bibliography{mainbib}
}

\end{document}